\documentclass[final,12pt]{colt2025_arxiv} % Include author names

\title[Optimality of roubly robust learning]{Structure-agnostic Optimality of Doubly Robust Learning for Treatment Effect Estimation}
\usepackage{times,verbatim,enumerate,amsfonts,bbm,mathabx,graphicx}
\usepackage{amsmath,amsfonts,bm}

\def\floor#1{\lfloor #1 \rfloor}
\def\1{\bm{1}}

\DeclareMathAlphabet{\mathsfit}{\encodingdefault}{\sfdefault}{m}{sl}
\SetMathAlphabet{\mathsfit}{bold}{\encodingdefault}{\sfdefault}{bx}{n}

\def\gC{{\mathcal{C}}}
\def\gD{{\mathcal{D}}}
\def\gE{{\mathcal{E}}}
\def\gF{{\mathcal{F}}}

\def\gH{{\mathcal{H}}}

\def\gO{{\mathcal{O}}}
\def\gP{{\mathcal{P}}}
\def\gQ{{\mathcal{Q}}}

\def\gS{{\mathcal{S}}}
\def\gT{{\mathcal{T}}}

\def\gX{{\mathcal{X}}}
\def\gY{{\mathcal{Y}}}

\newcommand{\E}{\mathbb{E}}

\newcommand{\R}{\mathbb{R}}

\newcommand\normx[1]{\left\Vert#1\right\Vert}
\newcommand\absx[1]{\left\vert#1\right\vert}
\newcommand{\att}{\textsc{ATT}}
\newcommand{\wate}{\textsc{WATE}}
\newcommand{\ate}{\textsc{ATE}}
\newcommand{\ml}{\textsc{ml}}
\newcommand{\cX}{\ensuremath{{\cal X}}}
\newcommand\ci{\protect\mathpalette{\protect\independenT}{\perp}}
\newtheorem{assumption}{Assumption}
\def\independenT#1#2{\mathrel{\rlap{$#1#2$}\mkern2mu{#1#2}}}
\usepackage{color-edits}
\addauthor{vs}{red}
\addauthor{jj}{blue}
\coltauthor{%
 \Name{Jikai Jin} \Email{jkjin@stanford.edu}\\
 \addr Stanford University
 \AND
 \Name{Vasilis Syrgkanis} \Email{vsyrgk@stanford.edu}\\
 \addr Stanford University%
}

\begin{document}

\maketitle

\begin{abstract}%
  Average treatment effect estimation is the most central problem in causal inference with application to numerous disciplines. While many estimation strategies have been proposed in the literature, the statistical optimality of these methods has still remained an open area of investigation, especially in regimes where these methods do not achieve parametric rates. In this paper, we adopt the recently introduced structure-agnostic framework of statistical lower bounds, which poses no structural properties on the nuisance functions other than access to black-box estimators that achieve some statistical estimation rate. This framework is particularly appealing when one is only willing to consider estimation strategies that use non-parametric regression and classification oracles as black-box sub-processes. Within this framework, we prove the statistical optimality of the celebrated and widely used doubly robust estimators for both the Average Treatment Effect (ATE) and the Average Treatment Effect on the Treated (ATT), as well as weighted variants of the former, which arise  in policy evaluation. \footnote{Accepted for presentation at the
Conference on Learning Theory (COLT) 2025.}
\end{abstract}


\section{Introduction}
\label{sec:intro}

Estimating the average treatment effect is one of the central problems in causal inference and has found important applications in numerous disciplines such as economics \citep{hirano2003efficient,imbens2004nonparametric}, education \citep{oreopoulos2006estimating}, epidemiology \citep{little2000causal,wood2008empirical} and political science \citep{mayer2011does}. In view of its practical importance, a large body of work is devoted to developing statistically efficient estimators for the average treatment effect based on regression \citep{robins1994estimation,robins1995analysis,imbens2003mean}, matching \citep{heckman1998matching,rosenbaum1989optimal,abadie2006large} and propensity scores \citep{rosenbaum1983central,hirano2003efficient} as well as their combinations. 

Despite the plethora of estimation algorithms for the average treatment effect, little is known about the statistical limits of estimating the average treatment effect within some formal minimax optimality framework. Existing minimax optimality results only apply to statistical quantities that resemble or are natural variants of the average treatment effect (see e.g. \citep{robins2009semiparametric,balakrishnan2019hypothesis,kennedy2022minimax,robins2008higher}). On the other hand, optimality results for the average effect are only known in the regime where the non-parametric components of the data generating process are estimable at a fast enough rate (typically $n^{-1/4}$). In this regime, the average effect is estimable at root-$n$ and the field of semi-parametric efficiency \citep{newey1994asymptotic} has provided optimal variance constants that multiply the leading rate. Finally, the prior work of \cite{bradic2019minimax}, characterizes minimax optimal conditions for root-$n$ estimability, albeit in a model where the effect is assumed to be constant for every unit in the population as well as other linearity assumptions. However, the optimal achievable estimation rate for any estimation quality of the non-parametric components has not been established. We provide the first tight statistical lower bound for the average treatment effect within the structure-agnostic minimax optimality framework \citep{balakrishnan2023fundamental}, which is an optimality framework that is a natural fit for understanding the limits of estimators that use machine learning algorithms as black-box regression oracles. Within this structure-agnostic paradigm, we show that the celebrated doubly robust estimation algorithm \citep{robins1994estimation} achieves minimax optimal mean-squared-error rates, up to constant factors.

Given a binary treatment $D\in \{0, 1\}$ and an outcome of interest $Y\in \R$, we let $Y(1), Y(0)$ denote the random potential outcomes that we would have observed from each unit, had we treated them with each possible treatment $d\in \{0,1\}$. Two central problems of causal analysis are the estimation of the \emph{average treatment effect} (\ate) and the \emph{average treatment effect on the treated} (\att) \citep{heckman1998matching}, defined correspondingly as the causal estimands:
\begin{align}
    \theta^{\ate} :=~& \E\left[Y(1) - Y(0)\right], &
    \theta^{\att} :=~& \E\left[Y(1) - Y(0)\mid D=1\right]
\end{align}
We consider the case when all potential confounders $X\in \cX \subseteq \R^K$ of the treatment and the outcome are observed; a setting that has received substantial attention in the causal inference literature. In particular, we will make the widely used assumption of \emph{conditional ignorability}: 
\begin{align}
    Y(1), Y(0) \ci D \mid X
\end{align}

We assume that we are given data that consist of samples of the tuple of random variables $(X, D, Y)$, that satisfy the basic \emph{consistency} property 
\begin{align}
    Y = Y(D)
\end{align} 
Without loss of generality, the data generating process obeys the regression equations:
\begin{equation}
    \label{model}
    \begin{aligned}
        Y =~& g_0(D,X) + U, & \E\left[ U\mid D,X\right]=~& 0\\
        D =~& m_0(X) + V, & \E\left[V\mid X\right]=~& 0
    \end{aligned}
\end{equation}
where $U,V$ are noise variables. The \emph{outcome regression} function $g_0(d,x)$ and the \emph{propensity score} $m_0(x)$ are commonly referred to as \textit{nuisance functions}. Note that when the outcome $Y$ is also binary, then the non-parametric functions $g_0$ and $m_0$, as well as the marginal probability law of the covariates $X$, fully determine the likelihood of the observed data.

Under conditional ignorability, consistency and the \emph{overlap assumption} that both treatment values are probable conditional on $X$, i.e., $m_0(X)\in [c, 1-c]$ almost surely, for some $c>0$, it is well known that the \ate~ and \att~ are identified by the statistical estimands:
\begin{align}
\theta^{\ate} =~& \E[g_0(1, X) - g_0(0, X)], &
\theta^{\att} =~& \E[Y - g_0(0, X) \mid D=1].
\end{align}
Our goal is to derive the statistically optimal estimation rates for the \ate~ and the \att. We will also be interested in a weighted variant of the average treatment effect (\wate):
\begin{align}
    \E[w(X)\, (Y(1) - Y(0))],
\end{align}
where $w(x)\in \R$ is a given weight function defined on $\cX$. Such weighted average effects typically arise in the evaluation of personalized policies, where $w: \cX\to \{0,1\}$ corresponds to a personalized treatment policy \citep{tao2019doubly,hirano2003efficient}. Note that the \ate~ is a special case of the \wate~ with $w(X)=1$. Similar to the \ate, the \wate~ is identified by the statistical estimand:
\begin{equation}
    \theta^{\wate} = \E\left[ w(X)\left(g_0(1,X)-g_0(0,X)\right) \right].
\end{equation}

Since the nuisance functions $g_0$ and $m_0$ in \eqref{model} are unknown and may have complex structures, and since the dimension $K$ of the covariates $X$ can be large relative to the number of data $n$ in many applications, it is extremely suitable to apply modern machine learning (ML) methods for the non-parametric, flexible and adaptive estimation of these nuisance functions, including penalized linear regression methods 
\citep{belloni2014pivotal,van2014asymptotically,chernozhukov2022automatic,zou2005regularization}, random forest methods \citep{breiman2001random,hastie2009random,biau2008consistency,wager2015adaptive,syrgkanis2020estimation}, gradient boosted forests \citep{friedman2001greedy,buhlmann2003boosting,zhang2005boosting} and neural networks \citep{schmidt2020nonparametric,farrell2021deep}, as well as ensemble and model selection approaches that combine all the above using out-of-sample cross-validation metrics \citep{wolpert1992stacked,zhang1993model,freund1997decision,van2007super,sill2009feature,wegkamp2003model,Arlot2010,chetverikov2021cross}. 

Motivated by the wide adoption and use of black-box adaptive estimation methods \citep{polley2019package,ledell2020h2o,wang2021flaml,karmaker2021automl} for these non-parametric components of the data generating process, as well as their superior empirical performance \citep{bach2024hyperparameter}, even in the context of treatment effect estimation, we will examine the problem of statistical optimality within the \emph{structure agnostic} minimax framework that was recently introduced in \citet{balakrishnan2023fundamental}. In particular, the only assumption that we will be making about our data generating process is that we have access to estimates $\hat{g}$ and $\hat{m}$ that achieve some statistical error rate, as measured by the mean-squared error, i.e.
\begin{equation}
    \notag
    \|\hat{g}(0, X) - g_0(0, X)\|_{P_X, 2} \leq e_n, 
    \|\hat{g}(1, X) - g_0(1, X)\|_{P_X, 2} \leq e_n',
    \|\hat{m}(X) - m_0(X)\|_{P_X, 2} \leq f_n,
\end{equation}
where for any function $v:\cX \to \R$, we denote $\|v(X)\|_{P_X,2}:=\sqrt{\E[v(X)^2]}$. Having access to such estimates for these two non-parametric components and imposing the aforementioned estimation error constraints on the data generating process, \emph{we resolve the optimal statistical rate achievable by any estimation algorithm for the parameters of interest.} 

The structure agnostic framework is particularly appealing as it essentially restricts any estimation approach to only use non-parametric regression estimates as a black-box and not tailor the estimation strategy to particular structural assumptions about the regression function or the propensity. These further structural assumptions can many times be brittle and violated in practice, rendering the tailored estimation strategy invalid or low-performing. Hence, the structure agnostic statistical lower bound framework has the benefit that it yields lower bounds that can be matched by estimation procedures that are easy to deploy and robust in their details. 

We show that up to constant factors no estimation algorithm for the \wate~ and \att~ can achieve squared error rates for the parameter of interest that are better than:
\begin{align}\label{eqn:lower-bounds}
    \Omega\left( \max\{e_n,e_n'\}\cdot f_n\cdot \left\Vert w\right\Vert_{P_X,\infty}^2 + \left\Vert w\right\Vert_{P_X,2}^2/n\right) ~~~~\text{and}~~~~ 
    \Omega\left( e_n\cdot f_n + 1/n\right)
\end{align}
respectively. These lower bounds apply even if we impose rate restrictions on stronger nuisance error metrics, e.g. $\|v(X)\|_{P_X,\infty} = \sup_{x\in \mathrm{supp}(P_X)} |v(x)|$. Furthermore, our lower bound constructions apply even when the outcome $Y$ is binary and, in the case of the \wate, they apply even when we know one of the two outcome response functions, i.e. $g_0(0, \cdot)$ or $g_0(1,\cdot)$. 
Importantly, these lower bounds are well-known to be achievable by the well-established and widely used doubly robust estimators derived from a first-order debiasing scheme, also known as estimators with the mixed bias property \citep{rotnitzky2021characterization}.

For general non-parametric functional estimation, it has been shown decades ago that if the function possesses certain smoothness properties, then higher-order debiasing schemes can be designed that lead to improved error rates \citep{bickel1988estimating,birge1995estimation}. Specifically, first-order debiasing methods are suboptimal even when the nuisance function estimators are minimax optimal. Estimators based on higher-order debiasing have also been proposed and analyzed for functionals that arise in causal inference problems \citep{robins2008higher,van2014higher,robins2017minimax,liu2017semiparametric,kennedy2022minimax}. However, the fast rates of these methods crucially rely on the structure of the underlying function classes. Unlike first-order debiasing methods, higher-order methods are \textit{not} structure-agnostic, in the sense that their error rates no longer apply to black-box estimators of the nuisance functions and the corresponding estimators are many times cumbersome to deploy in practice. 

In contrast, our results show that first-order debiasing is structure-agnostic optimal for estimating both WATE and ATT. Our results extend the recent work of \citet{balakrishnan2023fundamental}, which proposed the structure agnostic minimax optimality framework and proved the statistical optimality of doubly robust estimators of the expected conditional co-variance functional, defined as $\theta^{\textsc{Cov}} = \E[(D-\E[D\mid X])(Y - \E[Y\mid X])]$. However, the approach in \citet{balakrishnan2023fundamental} cannot be easily modified to handle the average treatment effect functionals that we study here and which arguably possess a more central role in the causal inference literature.

\subsection{Technical contributions}
\label{subsec:technical-contributions}

Our proof of the lower bounds uses the method of fuzzy hypotheses that reduces our estimation problem to the problem of testing a pair of \textit{mixtures} of hypotheses. Such methods are widely adopted in establishing lower bounds for non-parametric functional estimation problems \citep{tsybakov2008introduction} and have been used to address the minimax estimation errors of various causal functionals in different settings, including the expected conditional covariance \citep{robins2009semiparametric,balakrishnan2019hypothesis}, CATE function at a given point \citep{kennedy2022minimax} and variance-weighted ATE \citep{robins2008higher}. Surprisingly however, there is very little understanding of the estimation limit of the \emph{vanilla} ATE -- arguably a central parameter in causal inference -- even under Holder-smoothness assumptions that are extensively investigated in the literature. Focusing on the structure-agnostic setting introduced in the previous section, this paper takes an important step towards closing this gap.

Our main technical contribution is a collection of carefully-designed constructions of the hypotheses that are built on \textit{asymmetric perturbations} in the space of nuisance functions. We construct the perturbations in a sequential manner, with the perturbation of one nuisance estimate being dependent on the other nuisance. We note that due to the more complicated relationships between the estimand and the data distribution, existing ways to construct composite hypotheses \citep{robins2009semiparametric,kennedy2022minimax,balakrishnan2023fundamental} are no longer applicable to our setting, as we expand next.

In \citet{balakrishnan2023fundamental}, the authors investigate the estimation problem of three functionals: quadratic functionals in Gaussian sequence models, quadratic integral functionals and the expected conditional covariance. They establish their lower bound by reducing it to lower-bounding the error of a related hypothesis testing problem. The error is then lower-bounded by constructing priors (mixtures) of the composite null and alternate distribution. The priors they construct are based on adding or subtracting bump functions on top of a fixed hypothesis in a symmetric manner, which is a standard proof strategy for functional estimation problems \citep{ingster1994minimax,robins2009semiparametric,arias2018remember,balakrishnan2019hypothesis}. The reason why the proof strategy of \citet{balakrishnan2023fundamental} fails for \textsc{WATE} and \textsc{ATT} is that the functional relationships between the nuisance parameters and these target parameters take significantly different forms. Specifically, the target parameters that \citet{balakrishnan2023fundamental} investigates are all in the form of
\begin{equation}
    \label{eq:form}
    T(f,g)=\left\langle f,g\right\rangle_{\gH},
\end{equation}
where $f,g$ are unknown nuisance parameters that lie in some Hilbert space $\gH$. To be concrete, consider the example of the expected conditional covariance $\theta^{\textsc{Cov}}$. Let $\mu_0(x) = \E\left[ Y \mid X=x\right]$, then we have that $\theta^{\textsc{Cov}} = \E[DY] - \int m_0(x)\mu_0(x)\text{d} p_X(x)$
where $p_X$ is the marginal density of $X$. The first term, $\E[DY]$, can be estimated with a standard $\gO(n^{-{1}/{2}})$ rate, so what remains to be estimated is the second term which is exactly in the form of \eqref{eq:form}. However, the \ate~ and \att~ functionals do not take this inner product form. Instead they are, respectively, of the form:
\begin{align*}
    \notag
    T_1 (m_0,g_0) :=~& \E_{X} \left[g_0(1,X)-g_0(0,X) \right]= \E_{D,X} \left[\frac{D-m_0(X)}{m_0(X) (1-m_0(X))}g_0(D,X) \right]\\
    \notag
    T_2 (m_0,g_0) :=~& \frac{\E_X\left[\left(g_0(1,X)-g_0(0,X)\right)m_0(X)\right]}{\E_X\left[m_0(X) \right]}.
\end{align*}
Stepping outside of the realm of inner product functionals is the major challenge in extending existing approaches of establishing lower bounds to the problem of estimating \wate~ and \att, and very different constructions are required, which is our main technical innovation.

\subsection{Notation}
\label{subsec:notations}

We use $P_X$ to denote the marginal distribution of the confounding factors $X$ in the model (\ref{model}). For any function $f:\R^n\mapsto\R^k$ and distribution $P$ over $\R^n$, we define its $L^r$-norm as $\left\Vert f\right\Vert _{P,r} = \left(\int \left\Vert f\right\Vert^r \text{d}P \right)^{{1}/{r}},\quad r\in(0,+\infty)$
and $\left\Vert f\right\Vert_{P,\infty} = \text{ess sup} \left\{ f(X): X\sim P\right\}$.
We also slightly abuse notation and use $\left\Vert f\right\Vert _{r}$ instead, when the distribution is clear from context. For two sequences $(a_n)_{n\geq 1}$ and $(b_n)_{n\geq 1}$, we write $a_n=\gO(b_n)$ if there exists a constant $C>0$ such that $|a_n|\leq C|b_n|,\forall n\geq 1$, and we write $a_n=\Omega(b_n)$ if there exists a constant $c>0$ such that $|a_n|\geq c|b_n|,\forall n\geq 1$.

\section{Structure-agnostic estimation of average treatment effect}

To analyze the statistical limit of estimating treatment effect without making assumptions on regularity properties of nuisance functions, we adopt the structure-agnostic framework introduced by \cite{balakrishnan2023fundamental}. 
Specifically, we assume the existence of black-box estimates $\hat{m}(x)$ and $\hat{g}(d,x)$ of $m(x)$ and $g(d,x)$ that are accurate in the sense of $L^2$ distance:
\begin{equation}
    \label{eq:nuisance-accuracy}
    \begin{aligned}
        \left\Vert g_0(0,X)-\hat{g}(0,X)\right\Vert_{P_X,2}^2 \leq e_n,  \left\Vert g_0(1,X)-\hat{g}(1,X)\right\Vert_{P_X,2}^2 \leq~ e_n', 
        \left\Vert m_0(X)-\hat{m}(X)\right\Vert_{P_X,2}^2 \leq f_n,
    \end{aligned}
\end{equation}
where $e_n, e_n'$ and $f_n$ are \textit{arbitrary} positive numbers that depend on the sample size $n$ used to estimate the nuisance functions. Note that here we assume that the estimators $\hat{m},\hat{g}$ are already known to the statistician rather than a part of the estimation process. The reason for considering this setup is that we do not want to open the \textit{black box} of how these estimators are obtained. In practice, these estimators can be obtained by leveraging estimation methods such as Lasso \citep{bickel2009simultaneous}, random forest \citep{syrgkanis2020estimation}, deep neural networks \citep{chen1999improved,schmidt2020nonparametric,farrell2021deep} among others. 
% Ideally, we would like have a guarantee on the final estimation error that only depends on the nuisance estimation error (\ref{eq:nuisance-accuracy}) but not on algorithmic-dependent properties.
% Moreover, while we do not explicitly impose smoothness assumptions on the ground-truth nuisance functions $m_0$ and $g_0$, existing works that rely on such assumptions can still be related to our current setup, since the level of smoothness of the nuisance functions directly affects their minimax optimal estimation errors $e_n$ and $f_n$ \cite{kennedy2022minimax}.

Apart from the nuisance estimators, we also assume access to \emph{i.i.d.} data $\{(X_i,D_i,Y_i)\}_{i=1}^n$ that are also \emph{independent} of the data used to obtain the nuisance estimators. In this way, we fully disentangle the complete estimation procedure into a \textit{learning} phase where any estimation method can be used to obtain nuisance estimators from a portion of the data, and an second phase that leads to the final parameter estimate. While estimation of \textsc{WATE}/\textsc{ATT} does not necessarily need to follow this procedure, this is a typical pipeline implemented in practice, since it allows a flexible combination of black-box machine learning methods and estimators that cleverly leverage the structural properties of the model.
%We are interested in answering the following question: \textit{what is the optimal error rate that we can achieve for estimating \textsc{WATE} and \textsc{ATT}, given known estimators of nuisance functions and i.i.d. data $\{(X_i,D_i,Y_i)\}_{i=1}^n$?} 
As discussed before, the doubly robust estimators achieve error rates in the form of \eqref{eqn:lower-bounds}. Surprisingly, however, it has long been unknown whether one can actually do better than doubly robust estimators, which we address next.

\label{sec:upper-bound}

\section{Main results}
\label{sec:main-result}

In this section, we present our main results that lower-bound the estimation errors in the structural-agnostic setting. %Our lower bounds match the upper bounds derived in the previous section, implying that double/debiased ML estimators are structure-agnostic optimal in estimating \textsc{WATE} and \textsc{ATT}.

We restrict ourselves to the case of binary outcomes:

\begin{assumption}
    \label{asmp:outcome-binary}
    The outcome variable $Y$ is binary, \emph{i.e.}, $Y\in\{0,1\}$.
\end{assumption}

Given that the black-box nuisance function estimators satisfy \eqref{eq:nuisance-accuracy}, we define the following constraint set
\begin{equation}
    \label{local-set}
    \begin{aligned}
        \gF_{e_n,e_n',f_n} = \Bigl\{ & (m,g) \mid \mathrm{supp}(X)=[0,1]^K, P_X=\mathrm{Uniform}\bigl([0,1]^K\bigr),  \\
        & \Vert g(0,X)-\hat{g}(0,X)\Vert_{P_X,2}^2 \leq e_n, \Vert g(1,X)-\hat{g}(1,X)\Vert _{P_X,2}^2 \leq e_n', \\
        & \Vert m(X)-\hat{m}(X)\Vert _{P_X,2}^2 \leq f_n, 0\leq m(x),g(d,x)\leq 1,\forall x\in[0,1]^K, d\in\{0,1\}\Bigr\}
    \end{aligned}
\end{equation}
where
\begin{equation}
    \notag
    e_n, e_n', f_n = o(1)\quad (n\to +\infty).
\end{equation}

Note that introducing Assumption \ref{asmp:outcome-binary} and constraints on $P_X$ in \eqref{local-set} only strengthens the lower bound that we are going to prove, since they provide additional information on the ground-truth model. Moreover, the constraints $0\leq m(x), g(d,x)\leq 1$ naturally holds due to the fact that both the treatment and outcome variables are binary. We then define the minimax $(1-\gamma)$-quantile risk of estimating $\theta^{\wate}$ over a function space $\gF$ as
\begin{equation}
    \label{ATE-minimax-risk}
    \mathfrak{M}_{n,\gamma}^{\wate}\left(\gF\right) = \inf_{\hat{\theta}:\left(\gX\times\gD\times\gY\right)^n\mapsto\R} \sup_{(m^*,g^*)\in\gF} \gQ_{P_{m^*,g^*},1-\gamma}\left( \left|\hat{\theta}-\theta^{\wate}\right|^2 \right),
\end{equation}
where $\gQ_{P,\gamma}(X)=\inf\left\{x\in\R: P[X\leq x]\geq\gamma\right\}$ denotes the quantile function of a random variable $X$, and $P_{m^*,g^*}$ is the joint distribution of $\{(X_i,D_i,Y_i)\}_{i=1}^n$ which is uniquely determined by the functions $m^*$ and $g^*$. Specifically, let $\mu$ be the uniform distribution on $\gX\times\gD\times\gY=[0,1]^K\times\{0,1\}\times\{0,1\}$, then the density $p_{m^*,g^*}={\text{d}P_{m^*,g^*}}/{\text{d}\mu}$ can be expressed as
\begin{equation}
    \notag
    p_{m^*,g^*}(x,d,y) = m^*(x)^d(1-m^*(x))^{1-d}g^*(d,x)^y(1-g^*(d,x))^{1-y}.
\end{equation}

According to \eqref{ATE-minimax-risk}, $\mathfrak{M}_{n,\gamma}^{\wate}\left(\gF\right) \geq \rho$ would imply that for any estimator $\hat{\theta}$ of \textsc{WATE}, there must exist some $(m^*,g^*)\in\gF$, such that under the induced data distribution, the probability of $\hat{\theta}$ having estimation error $\geq\rho$ is at least $1-\gamma$. This provides a stronger form of lower bound compared with the minimax \emph{expected} risk defined in \cite{balakrishnan2023fundamental}, in the sense that the lower bound $\mathfrak{M}_{n,\gamma}^{\wate}\left(\gF\right) \geq \rho$ implies a lower bound $(1-\gamma)\rho$ of the minimax expected risk, but the converse does not necessarily hold.

Similarly, one can define the minimax quantile risk for estimating \textsc{ATT} as
\begin{equation}
    \label{ATTE-minimax-risk}
    \mathfrak{M}_{n,\gamma}^{\att}\left(\gF\right) = \inf_{\hat{\theta}:\left(\gX\times\gD\times\gY\right)^n\mapsto\R} \sup_{(m^*,g^*)\in\gF} \gQ_{P_{m^*,g^*},1-\gamma}\left( \left|\hat{\theta}\left(\left\{(X_i,D_i,Y_i)\right\}_{i=1}^n\right)-\theta^{\att}\right|^2 \right).
\end{equation}

The main objective of this section is to derive lower bounds for $\mathfrak{M}_{n,\gamma}^{\wate}\left(\gF_{e_n,e_n',f_n}\right)$ and $\mathfrak{M}_{n,\gamma}^{\att}\left(\gF_{e_n,e_n',f_n}\right)$ in terms of $e_n,e_n',f_n$ and $n$. We also need to assume that the estimators $\hat{m}(x): [0,1]^K\mapsto[0,1]$ and $\hat{g}(d,x):\{0,1\}\times[0,1]^K\mapsto[0,1]$ are bounded away from $0$ and $1$.

\begin{assumption}
\label{asmp:estimator-bounded}
    There is a constant $c\in(0,1/2)$ such that $c \leq \hat{m}(x), \hat{g}(d,x) \leq 1-c, \forall d\in\{0,1\}$, $x\in[0,1]^K$.
\end{assumption}

The assumption that $c \leq \hat{m}(x)\leq 1-c$ is common in deriving upper bounds for doubly robust estimators. On the other hand, the assumption that $c\leq \hat{g}(d,x)\leq 1-c$ is typically not needed for deriving upper bounds, but it is also made in prior works for proving lower bounds for estimating the expected conditional covariance $\E\left[\mathrm{Cov}(D,Y\mid X) \right]$ \citep{robins2009semiparametric,balakrishnan2023fundamental}. Now we are ready to state our main results.

\begin{theorem}
    \label{thm:ate}
    For any constant $\gamma\in\left({1}/{2},1\right)$ and estimators $\hat{m}(x)$ and $\hat{g}(d,x)$ that satisfy Assumption \ref{asmp:estimator-bounded}, for any given weight function $w$, the minimax risk of estimating the \textsc{WATE} is
    \begin{equation}
        \notag
        \mathfrak{M}_{n,\gamma}^{\wate}\left(\gF_{e_n,e_n',f_n}\right) = \Omega\left( \max\{e_n,e_n'\} f_n\cdot\Vert w\Vert _{P_X,\infty}^2 + \Vert w\Vert_{P_X,2}^2/n\right).
    \end{equation}
\end{theorem}

\begin{remark}
\label{remark:known-baseline}
    If we only assume that $c\leq\hat{m}(x), \hat{g}(1,x) \leq 1-c$ in Assumption \ref{asmp:estimator-bounded}, then we would have the lower bound
    \begin{equation}
        \notag
        \mathfrak{M}_{n,\gamma}^{\wate}\left(\gF_{e_n,e_n',f_n}\right) = \Omega\left( e_n' f_n\cdot\Vert w\Vert _{P_X,\infty}^2 + \Vert w\Vert_{P_X,2}^2/n\right).
    \end{equation}
    Furthermore, this lower bound still holds in the case where we know the baseline response, \emph{i.e.,} $\hat{g}(0,x)=g_0(0,x)=0$.
\end{remark}

\begin{theorem}
    \label{thm:atte}
    For any constant $\gamma\in\left({1}/{2},1\right)$ and estimators $\hat{m}(x)$ and $\hat{g}(d,x)$ that satisfy Assumption \ref{asmp:estimator-bounded}, the minimax risk of estimating the \textsc{ATT} is given by
    \begin{equation}
        \notag
        \mathfrak{M}_{n,\gamma}^{\att}\left(\gF_{e_n,e_n',f_n}\right) = \Omega\left( e_n f_n + 1/n\right).
    \end{equation}
\end{theorem}

\begin{remark}
    As discussed in Section \ref{subsec:prelim}, generic machine learning estimators are typically guaranteed to have small $L^2$ errors. However, the lower bounds presented in Theorem \ref{thm:ate} and \ref{thm:atte} still hold even if we replace the $L^2$ norm constraints in \eqref{local-set} are replaced with stronger $L^r (2\leq r\leq +\infty)$ constraints.
\end{remark}

Theorems \ref{thm:ate} and \ref{thm:atte} provide lower bounds of the minimax estimation errors for the \textsc{WATE} and \textsc{ATT}, in terms of the sample size and the estimation error of the black-box nuisance function estimators. Our lower bounds exactly matches the well-known upper bounds attained by the doubly robust estimators (see Section \ref{appsubsec:drl}), indicating that doubly robust estimators are minimax optimal in the structural-agnostic setup.

\section{Proof of Theorem \ref{thm:ate}}
\label{sec:proofs}

In this section, we give the proof outline of our main result, Theorem \ref{thm:ate}, for the lower bound of estimating \textsc{WATE}. Omitted details in the proof can be found in The proof of Theorem \ref{thm:atte} can be found in Section \ref{subsec:proof-atte} in the appendix. We first introduce some preliminary results that our proof will rely on. 

\subsection{Preliminaries}
\label{subsec:prelim}

%In this subsection, we introduce some known results that build the relationship between functional estimation and hypothesis testing, and then prove some preparatory results for the construction of hypotheses in subsequent sections. 
Let $H$ be the Hellinger distance defined as $H(P,Q) = \frac{1}{2}\int \big(\sqrt{P(\text{d}x)} - \sqrt{Q(\text{d}x)}\big)^2$
for any distributions $P,Q$. The first result that we will introduce is due to \cite{robins2009semiparametric} and upper-bounds the Hellinger distance between two mixtures of product measures.

Formally, let $\mathcal{X}=\cup_{j=1}^m \mathcal{X}_j$ be a measurable partition of the sample space. Given a vector $\lambda=\left(\lambda_1, \ldots, \lambda_m\right)$ in some product measurable space $\Lambda=\Lambda_1 \times \cdots \times \Lambda_m$, let $P$ and $Q_{\lambda}$ be probability measures on $\mathcal{X}$ such that the following statements hold:
\begin{enumerate}[1.]
    \item $P\left(\mathcal{X}_j\right)=Q_\lambda\left(\mathcal{X}_j\right)= p_j$ for every $\lambda \in \Lambda$, and
    \item The probability measures $P$ and $Q_{\lambda}$ restricted to $\mathcal{X}_j$ depend on the $j$-th coordinate $\lambda_j$ of $\lambda$ only.
\end{enumerate}

Let $p$ and $q_{\lambda}$ be the densities of the measures $P$ and $Q_\lambda$ that are jointly measurable in the parameter $\lambda$ and the observation $x$, and $\pi$ be a probability measure on $\Lambda$. Define $b= m\max _j \sup _\lambda \int_{\mathcal{X}_j} \left(q_\lambda-p\right)^2/p d \mu$
and the mixed density $q=\int q_\lambda d \pi(\lambda)$, then we have the following result.

\begin{lemma}
\label{semi-param-thm}
    (\cite{robins2009semiparametric}, Theorem 2.1, simplified version) Suppose that the mixed densities are equal, i.e. that $q=p$, and that $n\max\{1,b\}\max_j p_j \leq A$ for all $j$ for some positive constant $A$, then there exists a constant $C$ that depends only on $A$ such that, for any product probability measure $\pi=\pi_1 \otimes \cdots \otimes \pi_m$,
    $
    H\left(P^{\otimes n}, \int Q_\lambda^{\otimes n} d \pi(\lambda)\right) \leq \max_j p_j\cdot Cn^2b^2.
    $
\end{lemma}

\begin{remark}
    Theorem \ref{semi-param-thm} considers a special case of \cite{robins2009semiparametric}, Theorem 2.1. The original variant of the theorem considers a more general setting where the measures $p$ are also indexed by $\lambda$, i.e. $p_{\lambda}$ and where $p$ is the mixture density. Here, we only need the special cases where all $P_{\lambda}$'s are equal to $P$. The original version of the theorem also required that all $p_{\lambda}$ satisfy that $\underline{B} \leq p \leq \bar{B}$ for some constants $\underline{B}, \bar{B}$. In our special case, we no longer need to assume that. The only step in the proof of \cite{robins2009semiparametric} that makes use of this assumption is that $\max_j \sup_{\lambda} \int_{\gX_j} \frac{p^2}{p_{\lambda}}\frac{\text{d}\mu}{p_j} \leq \frac{\bar{B}}{\underline{B}}$
    (see the arguments following their proof of Lemma 5.2). However, in our setting this term is simply
    \begin{equation}
        \notag
        \max_j \sup_{\lambda} \int_{\gX_j} \frac{p^2}{p_{\lambda}}\frac{\text{d}\mu}{p_j} = \max_j p_j^{-1} \int_{\gX_j} p \text{d}\mu = \max_j p_j^{-1} P(\gX_j) = 1.
    \end{equation}
\end{remark}

\begin{lemma}
\label{fano-method}
    (\cite{tsybakov2008introduction}, Theorem 2.15) Let $\pi$ be a probability distribution on a set (measure space) of distributions $\mathcal{P}$ with common support $\gX$, which induce the distribution
    $
    Q_1(A)=\int Q^{\otimes n}(A) d \pi(Q), \quad \forall A \subset \gP.
    $
    Suppose that there exists $P\in\gP$ and a functional $T: \gP\mapsto\R$ which satisfies 
    \begin{equation}
        \label{fano:separation-condition}
        T(P)\leq c, \quad \pi(\{Q: T(Q) \geq c+2 s\})=1
    \end{equation}
    for some $s>0$. If $H^2\left(P^{\otimes n}, Q_1\right) \leq \delta<2$, then $\inf_{\hat{T}: \gX^n\mapsto\R} \sup_{P \in \mathcal{P}} P\left[\left|\hat{T}-T(P)\right|\geq s\right] \geq {\big(1-\sqrt{\delta(1-\delta / 4)}\big)}/{2}.$
    Then  it holds that $\inf_{\hat{T}: \gX^n\mapsto\R} \sup_{P \in \mathcal{P}} \gQ_{P,1-\gamma}\left(\left|\hat{T}-T(P)\right|^2\right) \geq s^2$, where $\gamma = {\big(1+\sqrt{\delta(1-\delta / 4)}\big)}/{2}$, 
\end{lemma}

\subsection{Partitioning the covariate space} The following lemma states that for an arbitrary weight function $w(x)$, one can always partition the domain into two subsets that have the same amount of weights.

\begin{definition}
    We say that a set $\gS\subseteq\R^K$ is a hyperrectangle collection if it can be partitioned into a finite number of disjoint hyperrectangles in $\R^K$.
\end{definition}

\begin{lemma}
    \label{lemma:divide-half}
    Let $\gS\subseteq\R^K$ be a hyperrectangle collection and $w(x): [0,1]^K\mapsto\R$ be a non-negative Lebesgue-integrable function such that $\int_{[0,1]^K} w(x) \text{d} \mu_L(x) > 0$,
    then $\gS$ can be partitioned into two hyperrectangle collections $\gS_1, \gS_2$ such that $\mu_L(\gS_1) = {\mu_L(\gS)}/2$ and
    \begin{equation}
        \notag
        \int_{\gS_1} w(x) \text{d} \mu_L(x) = \frac{1}{2} \int_{\gS} w(x) \text{d} \mu_L(x),
    \end{equation}
    where $\mu_L$ is the Lebesgue measure on $\R^K$.
\end{lemma}

Let $P_X$ be the uniform distribution on $\mathrm{supp}(X)=[0,1]^K$ and $p_X$ be its density. It is easy to see that $P_X\big[|w(X)|>\Vert w\Vert _{P_X, \infty}/2\big] > 0$. Assuming without loss of generality that $P_X\big[w(X)>\Vert w\Vert _{P_X, \infty}/2\big] > 0$ (otherwise we consider $-w$ instead of $w$), we can define the "truncated" weight function $\hat{w}(x)=w(x)\mathbbm{1}\left\{w(x)>\Vert w\Vert_{P_X, \infty}/2\right\}$. Applying Lemma \ref{lemma:divide-half} to $\hat{w}\cdot w$, recursively, for $m$ times, with $m\in\mathbb{Z}_{+}$, we can partition $[0,1]^K$ into $M=2^m$ hyperrectangle collections $B_1, B_2,\cdots, B_M$, such that $\mu_L(B_j) = 1/M$ and
\begin{equation}
    \notag
    \int_{B_j} w(x)\hat{w}(x) \text{d} x = \frac{1}{M}\int_{[0,1]^K} w(x)\hat{w}(x) \text{d} x,\quad j=1,2,\cdots,M.
\end{equation}
Since $P_X$ is the uniform distribution on $[0,1]^K$, the above implies that
\begin{equation}
    \notag
    \E_{X}\left[ w(X)\hat{w}(X)\mathbbm{1}\{X\in B_j\} \right] = \frac{1}{M}\int_{[0,1]^K} w(x)\hat{w}(x) \text{d} x,\quad j=1,2,\cdots,M.
\end{equation}

Let $\lambda_i, i=1,2,\cdots, M$ be i.i.d. Rademacher random variables taking values $+1$ and $-1$ both with probability $0.5$. We define 
\begin{equation}
    \label{eq:def-Delta}
    \Delta(\lambda,x) = \sum_{j=1}^{M/2} \lambda_j\left(\mathbbm{1}\left\{x\in B_{2j}\right\}-\mathbbm{1}\left\{x\in B_{2j-1}\right\}\right).
\end{equation}

%The following properties of $\Delta(\lambda,x)$ are straightforward.

\begin{proposition}
\label{prop:delta-prop}
    We have
    \begin{subequations}
        \label{eq:Delta-properties}
        \begin{align}
            &\E_{\lambda} \Delta(\lambda,x) = \sum_{j=1}^{M/2} \E\lambda_j\left(\mathbbm{1}\left\{x\in B_{2j}\right\}-\mathbbm{1}\left\{x\in B_{2j-1}\right\}\right) = 0,\quad\forall x\in[0,1]^K \label{eq:Delta-prop-1} \\
            &\E_X w(X)\hat{w}(X) \Delta(\lambda,X) 
                = 0,\quad\forall\lambda\in\{0,1\}^{M/2}
             \label{eq:Delta-prop-2} \\
            &\Delta(\lambda,x)^2 = \sum_{j=1}^{M/2} \left(\mathbbm{1}\left\{x\in B_{2j}\right\}-\mathbbm{1}\left\{x\in B_{2j-1}\right\}\right)^2 = 1,\quad\forall x\in[0,1]^K,\lambda\in\{0,1\}^{M/2}. \label{eq:Delta-prop-3}
        \end{align}
    \end{subequations}
\end{proposition}

\begin{remark}
The construction of \emph{bump functions} $\Delta(\lambda,x)$ in the form of \eqref{eq:def-Delta} has also been used in a line of prior works for proving minimax lower bounds \cite{balakrishnan2023fundamental}. However, here we need to carefully construct the partition $B_j$ of the whole domain to handle non-uniform weights. We note that if we only wanted to deal with an ATE and not a WATE, then we would have simply chosen $B_j$ to be an equi-partition of the $[0,1]^K$ space and the above constructions of the regions $B_j$, related to balancing the given weights, would not be needed.
\end{remark}

\subsection{Core part of lower bound construction} Having completed all preparation steps, we are now ready to present our proof for Theorem \ref{thm:ate}. The remaining part of Section \ref{sec:proofs} is organized as follows. In Section \ref{subsec:case1} and \ref{subsec:case2}, we first establish our lower bound $\Omega\left(e_n'f_n\Vert w\Vert _{P_X,\infty}^2\right)$ under the following weaker version of Assumption \ref{asmp:estimator-bounded}, as previously mentioned in Remark \ref{remark:known-baseline}:

\begin{assumption}\label{assm:one-side-overlap}
    There exists a constant $c>0$ such that $c \leq \hat{m}(x), \hat{g}(1,x) \leq 1-c, \forall x\in[0,1]^K$.
\end{assumption}

We separately present our proof of this lower bound for the two cases $e_n'\geq f_n$ and $e_n'<f_n$. Interestingly, these two cases need to be handled separately using different constructions of the composite hypotheses. In Section S.2 in the supplementary material, we show how the lower bound $\Omega\left(e_nf_n\Vert w\Vert _{P_X,\infty}^2\right)$ can be derived in a similar fashion. To conclude our proof, it remains to prove the lower bound $\gO\left(n^{-1}\Vert w\Vert _{P_X,2}^2\right)$, which is the standard oracle error and can be found in the supplementary material. 

\subsection{Case 1: $e_n\geq f_n$}
\label{subsec:case1}
In this case, we define
\begin{equation}
    \label{eq:ate-construction-1}
    \begin{aligned}
        & g_{\lambda}(0,x) = \hat{g}(0,x), \quad g_{\lambda}(1,x) = \frac{\hat{m}(x)}{m_{\lambda}(x)} \left[\hat{g}(1,x) + \alpha\hat{w}(x)\Delta(\lambda,x) \right], \\
        & m_{\lambda}(x) = \hat{m}(x)\left[ 1 -  \frac{\beta}{\hat{g}(1,x)}\hat{w}(x)\Delta(\lambda,x)\right] 
    \end{aligned}
\end{equation}
where $\alpha,\beta>0$ are constants that will be specified later in Lemma \ref{lemma:ate-case1-bounds-collect}, where we will verify that $(m_{\lambda},g_{\lambda})$ belongs to the constrained set $\gF_{e_n,e_n',f_n}$ and thus are valid probabilities in particular. Compared with standard approaches for constructing the composite hypotheses \cite{ingster1994minimax,robins2009semiparametric,arias2018remember,balakrishnan2019hypothesis}, we employ an \emph{asymmetric} construction which means that the nuisance functions are \emph{non-linear} in the Rademacher variables $\lambda$ (in particular the function $g_{\lambda}$ depends non-linearly in $\lambda$ due to the dependence on $m_{\lambda}$ in the denominator). As discussed in Section \ref{sec:intro}, such type of non-standard constructions are necessary since the functional that we need to estimate has a different structure than those handled in previous works. We first prove some basic properties of our construction.

\begin{proposition}
\label{prop:lambda-avg-1}
    For all $x\in[0,1]^K$, we have
    \begin{subequations}
        \label{eq:lambda-avg-1}
        \begin{align}
            &\E_{\lambda} m_{\lambda}(x) = \hat{m}(x) - \hat{m}(x)\frac{\beta}{\hat{g}(1,x)}\hat{w}(x)\E_{\lambda}\Delta(\lambda,x) = \hat{m}(x) \label{eq:lambda-avg-1-m} \\
            &\E_{\lambda} \left[g_{\lambda}(1,x)m_{\lambda}(x)\right] = \hat{m}(x)\left(\hat{g}(1,x)+\alpha\hat{w}(x)\E_{\lambda}\Delta(\lambda,x)\right) = \hat{g}(1,x)\hat{m}(x). \label{eq:lambda-avg-1-gm}
        \end{align}
    \end{subequations}
\end{proposition}

We start by bounding the $L^2$ distance from $g_{\lambda}, m_{\lambda}$ to $\hat{g},\hat{m}$.

\begin{lemma}
\label{lemma:nuisance-radius-1}
    Assuming that $\beta\leq{c\normx{w}_{P_X,\infty}^{-1}}/2$ where $c$ is the constant introduced in Assumption \ref{asmp:estimator-bounded}, then the following holds for all $0<r\leq +\infty$:
    \begin{equation}
        \notag
        \begin{aligned}
            \normx{g_{\lambda}(1,X)-\hat{g}(1,X)}_{P_X,r} &\leq 2(\alpha+c^{-1}\beta)\Vert \hat{w}(X)\Vert _{P_X,r},\\
            \normx{m_{\lambda}(X)-\hat{m}(X)}_{P_X,r} &\leq c^{-1}\beta\Vert \hat{w}(X)\Vert _{P_X,r}.
        \end{aligned}
    \end{equation}
\end{lemma}

Let $Q_{\lambda}$ be the joint distribution of $(X,D,Y)$ induced by $g_{\lambda}$ and $m_{\lambda}$ and $\mu$ be the uniform distribution on $[0,1]^K\times\{0,1\}\times\{0,1\}$. Define $q_{\lambda} = {\text{d} Q_{\lambda}}/{\text{d} \mu}$.
Similarly, let $\hat{P}$ be the joint distribution of $(X,D,Y)$ induced by $\hat{g}$ and $\hat{m}$, and $\hat{p}={\text{d} \hat{P}}/{\text{d}\mu}$. The next lemma states that the mixture of $Q_{\lambda}$ with prior $\pi(\lambda)$ is exactly equal to $\hat{P}$.

\begin{lemma}
    \label{lemma:ate-case1-mixture-equal}
    Let $Q = \int Q_{\lambda} \text{d} \pi(\lambda)$ and $q = {\text{d} Q}/{\text{d} \mu} = \int q_{\lambda} \text{d} \pi(\lambda)$, then $\hat{p}=q$.
\end{lemma}

The following lemma implies that the  Hellinger distance between the empirical distribution under $\hat{P}$ and $Q_{\lambda}$ with prior $\pi(\lambda)$ can be made arbitrarily small, as long as the domain $\mathrm{supp}(X)$ is partitioned into sufficiently many pieces.

\begin{lemma}
\label{lemma:ate-case1-hellinger-bound}
    For any $\delta>0$, as long as $M \geq \max\{n,32Cn^2/(c^4\delta)\}$ where $c$ is the constant introduced in Assumption~\ref{assm:one-side-overlap} and $C$ is the constant implied by Lemma~\ref{semi-param-thm} for $A=4c^{-2}$, we have $H^2\left(\hat{P}^{\otimes n},\int Q_{\lambda}^{\otimes n}\text{d} \pi(\lambda)\right) \leq \delta.$
\end{lemma}

As the final building block for establishing our lower bound, we prove the following lemma, which implies that with proper choices of $\alpha$ and $\beta$, $m_{\lambda}, g_{\lambda}$ are close (in the sense of $L^2$-distance) to $\hat{m}$ and $\hat{g}$ respectively, and that the separation condition (\ref{fano:separation-condition}) holds with distance $s=\Omega\left(\sqrt{e_n f_n}\normx{w}_{P_X,\infty}\right)$. 

\begin{lemma}
\label{lemma:ate-case1-bounds-collect}
    Let $\alpha = \sqrt{e_n'}/\big(4\Vert \hat{w}(X)\Vert _{P_X,2}\big),\quad \beta = c\sqrt{f_n}/\big(4\Vert \hat{w}(X)\Vert _{P_X,2}\big)$, 
    then for sufficiently large $n$, we have $(m_{\lambda}, g_{\lambda})\in\gF_{e_n,e_n',f_n}$ and
    \begin{equation}
        \label{eq:list-cond-1.3}
        \E_X\left[w(X) g_{\lambda}(1,X)\right] \geq \E\left[w(X) \hat{g}(1,X) \right] + \frac{1}{2}\alpha\beta\E\left[\frac{w(X)\hat{w}(X)^2}{\hat{g}(1,X)}\right],\forall\lambda\in\{0,1\}^{M/2}.
    \end{equation}
\end{lemma}

We are now ready to prove Theorem \ref{thm:ate} in the case when $e_n'\geq f_n$. For any $\gamma>{1}/{2}$, there exists some $\delta\in(0,2)$ such that ${(1+\sqrt{\delta(1-\delta/4)})}/{2}=\gamma$. We choose $M \geq \max\{n,32Cn^2/{c^4\delta}\}$ and $\gP=\{\hat{P}\}\cup\left\{Q_{\lambda}:\lambda\in\{0,1\}^{M/2}\right\}$, $P=\hat{P}$, $\pi$ be the discrete uniform distribution on $\left\{Q_{\lambda}:\lambda\in\{0,1\}^{M/2}\right\}$, $s=\frac{1}{4}\alpha\beta \E\big[{w(X)\hat{w}(X)^2}/{\hat{g}(1,X)}\big]$ in the context of Lemma \ref{fano-method}. Then Lemma \ref{lemma:ate-case1-hellinger-bound} and \ref{lemma:ate-case1-bounds-collect} imply that all the listed conditions are satisfied for the \textsc{WATE} functional $T(P) = \theta^{\wate}(P) = \E_P\left[ w(X)\left( g(1,X)-g(0,X)\right)\right]$.
Therefore, by Lemma \ref{fano-method}, we have 
\begin{equation}
    \notag
    \begin{aligned}
        &\quad \inf_{\hat{\theta}}\sup_{P\in\gP} \gQ_{P,1-\gamma}\left(\left| \hat{\theta}\left(\{(X_i,D_i,Y_i)\}_{i=1}^N\right) - \theta^{\wate}\right|^2\right) = \Omega\left(\alpha\beta \E\left[\frac{w(X)\hat{w}(X)^2}{\hat{g}(1,X)}\right]\right) \\
        &= \Omega\left( \frac{\sqrt{e_n' f_n}}{\Vert \hat{w}(X)\Vert _{P_X,2}^2}\cdot \E\left[\frac{w(X)\hat{w}(X)^2}{\hat{g}(1,X)}\right]\right) = \Omega\left( \sqrt{e_n' f_n}\cdot \frac{\E\left[w(X)\hat{w}(X)^2\right]}{\Vert \hat{w}(X)\Vert _{P_X,2}^2}\right) \\
        &= \Omega\left( \sqrt{e_n' f_n}\cdot \frac{\E\left[ w(X)^3\mathbbm{1}\left\{ w(X) > \frac{1}{2}\Vert w\Vert _{P_X,\infty} \right\} \right]}{\E\left[ w(X)^2\mathbbm{1}\left\{ w(X) > \frac{1}{2}\Vert w\Vert _{P_X,\infty} \right\} \right]}\right) = \Omega\left( \Vert w\Vert _{P_X,\infty}\sqrt{e_n' f_n}\right).
    \end{aligned}
\end{equation}

\subsection{Case 2: $f_n > e_n'$}
\label{subsec:case2}
In this case, we consider a different construction as follows:
\begin{equation}
    \label{eq:ate-construction-2}
    \begin{aligned}
        & g_{\lambda}(0,x) = \hat{g}(0,x), \quad g_{\lambda}(1,x) = \frac{\hat{g}(1,x)}{1+\frac{\beta}{\hat{g}(1,x)}\hat{w}(x)\Delta(\lambda,x)-\alpha\beta\hat{w}(x)^2}, \\
        & m_{\lambda}(x) = \frac{\hat{g}(1,x)}{g_{\lambda}(1,x)}\left(\hat{m}(x) + \alpha\hat{m}(x)\hat{g}(1,x)\hat{w}(x)\Delta(\lambda,x)\right)
    \end{aligned}
\end{equation}
where $\Delta(\lambda,x)$ is defined in \eqref{eq:def-Delta} and $\alpha,\beta>0$ are constants that need to be specified later. The remaining steps follow a similar reasoning as the previous subsection. Due to space limit, we present the proof in Section \ref{appsubsec:case2-complete}.

\section{Conclusion}

We investigated the statistical limit of treatment effect estimation in the structural-agnostic regime, which is an appropriate lower bound technique when one wants to only consider estimation strategies that use generic black-box estimators for the various nuisance functions involved in the estimation of treatment effects. We establish the minimax optimality of the celebrated and widely used in practice doubly robust learning strategies via reducing the estimation problem to a hypothesis testing problem, and lower bound the error of any estimation algorithm via non-standard constructions of the fuzzy hypotheses. Our results show that these estimators are optimal, in the structure agnostic sense, even in the slow rate regimes, where the implied rate for the functional of interest is slower than root-$n$. Hence, any improvements upon these estimation strategies need to incorporate elements of the structure of the nuisance functions and cannot simply invoke generic adaptive regression approaches as black-box sub-processes. While the focus of this paper is on treatment effect estimation, we believe that our techniques can be extended to address structure agnostic minimax lower bounds of more general functional estimation problems.

% Acknowledgments---Will not appear in anonymized version
\acks{VS is supported by NSF Award IIS-2337916. JJ is partially supported by NSF Award IIS-2337916.}

\bibliography{example}

\appendix

% \crefalias{section}{appendix} % uncomment if you are using cleveref
\newpage
\section{Background on doubly robust learning}
\label{appsec:drl}
\subsection{Doubly robust estimators for treatment effect estimation}
\label{appsubsec:drl}

If we have access to estimates $\hat{g}$ and $\hat{m}$, a straightforward approach to estimating our target quantities is to directly plug these estimators in the formulas that describe our statistical estimands. 
\begin{comment}
i.e., 
\begin{align}
    \hat{\theta}_{\textrm{plugin}}^{\wate} =~& \frac{1}{n}\sum_{i=1}^n w(X_i) \left(\hat{g}(1,X_i)-\hat{g}(0,X_i)\right), 
    \label{eq:ate-plug-in}\\
    \hat{\theta}_{\textrm{plugin}}^{\att} =~& \left(\sum_{i=1}^n D_i\right)^{-1}\sum_{i=1}^n D_i\, \left(Y_i -\hat{g}(0,X_i)\right). \label{eq:ATT-plug-in}
\end{align}
\end{comment}
This approach renders the estimation accuracy of the target parameter very susceptible to estimation errors of the outcome regression nuisance function, which could be large due to high-dimensionality, regularization and model selection. Moreover, the function spaces over which these estimators operate might not be simple and do not necessarily satisfy a widely invoked Donsker condition \citep{dudley2014uniform}. 

To mitigate this heavy dependence on the outcome regression model and to lift any restrictions on the form of the nuisance estimation algorithm, other than mean-squared-error accuracy, a line of recent works \citep{chernozhukov2017double,chernozhukov2018double,foster2023orthogonal,rotnitzky2021characterization,chernozhukov2022locally,chernozhukov2023simple} proposes the use of sample splitting, together with first-order debiasing correction approaches that lead to estimating equations that satisfy the property of Neyman orthogonality. Several ideas in this line of work have also been explored in the strongly related variant of targeted learning \citep{van2006targeted,van2011cross} and derive inspiration from the earlier classical work of \citet{bickel1982adaptive,schick1986asymptotically,klaassen1987consistent,robinson1988root,bickel1993efficient,goldstein1992optimal,newey1994asymptotic,ai2003efficient} in the field of semi-parametric inference. These semi-parametric estimators attain root-$n$ rates for the parameter of interest assuming that the non-parametric nuisance estimates attain mean-squared-error rates that decay faster than $n^{1/4}$ and impose no further restrictions on the function spaces used in estimation or any other properties that the nuisance estimators need to satisfy.

In the case of average treatment effect estimation this approach leads to a sample-splitting variant of the well-known doubly robust estimators \citep{robins1995analysis,robins1995semiparametric} of the $\wate$ and the $\att$, i.e.:
\begin{align}
        \label{eq:ate-debias-estimator}
        \hat{\theta}^{\wate} = \frac{1}{n}\sum_{i=1}^n w(X_i)\left[ \hat{g}(1,X_i) - \hat{g}(0,X_i) + \frac{D_i - \hat{m}(X_i)}{\hat{m}(X_i)\,(1-\hat{m}(X_i))}\left( Y_i - \hat{g}(D_i,X_i)\right) \right]\\
        \label{ATT-debias-estimator}
        \hat{\theta}^{\att} = \left(\sum_{i=1}^n D_i\right)^{-1} \sum_{i=1}^n \left[ D_i\left(Y_i-\hat{g}(0,X_i)\right) - \frac{\hat{m}(X_i)}{1-\hat{m}(X_i)}(1-D_i)(Y_i-\hat{g}(0,X_i))\right]
\end{align}

Even though the $n^{1/4}$ rate requirement can be achieved by a broad range of machine learning methods \citep{bickel2009simultaneous,belloni2011l1,belloni2013least,chen1999improved,wager2018estimation,athey2019generalized} (under assumptions), it can many times be violated in practice. Even in the case when this requirement is violated a small modification of the arguments employed in \citet{chernozhukov2018double,foster2023orthogonal} can be invoked to prove the structure-agnostic upper bounds stated below.

\begin{theorem}
\label{thm:ate-upper-bound}\label{thm:ATT-upper-bound}
    Suppose that there exists a constant $c\in(0,1)$ such that $c\leq \hat{m}(x)\leq 1-c, \forall x\in\mathrm{supp}(X)$ and $|Y|\leq G$ \emph{a.s.}, for some constant $G$. Then for any $\delta>0$, there exists a constant $C_{\delta}$ such that the doubly robust estimator of the \textsc{WATE} (defined in \eqref{eq:ate-debias-estimator})
    achieves estimation error
    \begin{equation}
        \label{eq:ate-upper-bound}
        \left|\hat{\theta}^{\wate}-\theta^{\wate}\right|^2 \leq C_{\delta}\left( \max\{e_n,e_n'\}\cdot f_n\cdot \left\Vert w\right\Vert_{P_X,\infty}^2 + \frac{1}{n}\left\Vert w\right\Vert_{P_X,2}^2\right).
    \end{equation}
    with probability $\geq 1-\delta$. Moreover, the doubly robust estimator of the \textsc{ATT} (defined in \eqref{ATT-debias-estimator})
    achieves estimation error
    \begin{equation}
        \label{eq:att-upper-bound}
        \left|\hat{\theta}^{\att}-\theta^{\att}\right|^2 \leq C_{\delta}\left(e_n\cdot f_n+\frac{1}{n}\right)
    \end{equation}
    with probability $\geq 1-\delta$.
\end{theorem}

Theorem \ref{thm:ate-upper-bound} implies that with high probability, the estimation error of the debiased estimator (\ref{eq:ate-debias-estimator}) is upper-bounded by the sum of the oracle error which equals ${1}/{n}$ multiplied by the $L^2$ norm of weight function $w$, and the product of the error in estimating nuisance functions $m_0$ and $g_0$, multiplied by the $L^{\infty}$ norm of $w$. Similarly, for estimating the \textsc{ATT}, Theorem \ref{thm:ATT-upper-bound} implies that with high probability, the error of the doubly robust estimator (\ref{ATT-debias-estimator}) is upper-bounded by the sum of the oracle error ${1}/{n}$ and the product of the error in estimating nuisance functions $m_0$ and $g_0(0,\cdot)$. 

\subsection{Proof of Theorem \ref{thm:ate-upper-bound}}

We define
\begin{equation}
    \notag
    \bar{\theta}^{WATE} = \E w(X)\left[ \hat{g}(1,X) - \hat{g}(0,X) + \left(\frac{D}{\hat{m}(X)}- \frac{1-D}{1-\hat{m}(X)}\right) \left( Y - \hat{g}(D,X)\right) \right],
\end{equation}
then $\E\hat{\theta}^{WATE}=\bar{\theta}^{WATE}$, which implies that
\begin{equation}
    \notag
    \E\left(\hat{\theta}^{WATE}-\bar{\theta}^{WATE}\right)^2 \leq \frac{1}{n} \mathrm{Var}\left(\hat{\theta}^{WATE}\right) \lesssim \frac{1}{n}\Vert w\Vert_{P_X,2}^2.
\end{equation}
On the other hand,
\begin{equation}
    \notag
    \begin{aligned}
        &\quad \left|\theta^{WATE}-\bar{\theta}^{WATE}\right| \\
        &\leq \E w(X) \left|1-\frac{m_0(X)}{\hat{m}(X)}\right|\left|g_0(1,X)-\hat{g}(1,X)\right| + \E w(X) \left|1-\frac{1-m_0(X)}{1-\hat{m}(X)}\right|\left|g_0(0,X)-\hat{g}(0,X)\right| \\
        &\leq \|w\|_{\infty}\cdot \left( \E \left|1-\frac{m_0(X)}{\hat{m}(X)}\right|\left|g_0(1,X)-\hat{g}(1,X)\right| + \E \left|1-\frac{1-m_0(X)}{1-\hat{m}(X)}\right|\left|g_0(0,X)-\hat{g}(0,X)\right| \right) \\
        &\leq c^{-1}\|w\|_{\infty}\|m_0(X)-\hat{m}(X)\|_{P_X,2}\cdot\left(\left\|g_0(0,X)-\hat{g}(0,X)\right\|_{P_X,2}+ \left\|g_0(1,X)-\hat{g}(1,X)\right\|_{P_X,2} \right) \\
        &= \gO\left( \|w\|_{\infty}\sqrt{\max\{e_n,e_n'\}f_n}\right).
    \end{aligned}
\end{equation}
Combining the above inequalities, we have
\begin{equation}
    \notag
    \E \left(\hat{\theta}^{WATE}-\theta^{WATE}\right)^2 = \mathcal{O}\left( \max\{e_n,e_n'\}\cdot f_n\cdot \|w\|_{\infty}^2 + \frac{1}{n}\right)
\end{equation}
and the desired high-probability bound follows directly from Markov's inequality.

Since $\E[D]=\E_X[m_0(X)]$ and $D_i,i=1,2,\cdots,n$ are i.i.d. Bernoulli variables, by central limit theorem there exists constant $\tilde{C}_{\delta,1}>0$ such that
\begin{equation}
    \label{eq:clt}
    \left|\frac{1}{n}\sum_{i=1}^n D_i - \E[D_1]\right| \leq C_{\delta,1}\sqrt{\frac{\mathrm{Var}(D_1)}{n}}\quad \text{with probability } \geq 1-\frac{1}{2}\delta.
\end{equation}
Hence with probability $\geq 1-\delta$, we have
\begin{equation}
    \notag
    \begin{aligned}
        &\quad \left|\hat{\theta}^{ATT}-\theta^{ATT}\right| \\
        &\lesssim \frac{1}{\sqrt{n}} + (\E[m_0(X)])^{-1}\left|(\E_n-\E)\left(D\left(Y-\hat{g}(0,X)\right) - \frac{\hat{m}(X)}{1-\hat{m}(X)}(1-D)(Y-\hat{g}(0,X))\right)\right| \\
        &\quad + \E\Big| m_0(X)\left(g_0(1,X)-g_0(0,X)\right) - m_0(X)\left(g_0(1,X)-\hat{g}(0,X)\right) \\
        &\quad -\hat{m}(X)\frac{1-m_0(X)}{1-\hat{m}(X)}\left(g_0(0,X)-\hat{g}(0,X)\right)  \Big| \\
        &\lesssim\frac{1}{\sqrt{n}} + \E\Big| \frac{(m_0(X)-\hat{m}(X))(g_0(0,X)-\hat{g}(0,X))}{1-\hat{m}(X)}\Big| \lesssim \frac{1}{\sqrt{n}} + \sqrt{e_nf_n},
    \end{aligned}
\end{equation}
where $\E_n$ denotes the empirical average in the second term of the second line, and this term is bounded by $\gO\left(\frac{1}{\sqrt{n}}\right)$ with high probability. This concludes the proof.

\section{Technical details for proving Theorem \ref{thm:ate}}

\subsection{Omitted proofs in Section \ref{sec:proofs}}
\label{appsubsec:omitted-proofs}

\subsubsection{Proof of Lemma \ref{lemma:divide-half}}

Suppose that $\gS = \cup_{i=1}^n \gC_i$, where $\gC_i = \bigtimes_{j=1}^K [a_{ij}, b_{ij}]$ are disjoint hyperrectangles. Let 
    \begin{equation}
        \notag
        \gT_{\alpha} = \bigcup_{i=1}^n \left(\bigtimes_{j=1}^{K-1}[a_{ij}, b_{ij}] \times \left[ \left(1-\frac{\alpha}{2}\right)\,a_{iK}+ \frac{\alpha}{2}b_{iK}, \frac{1-\alpha}{2}a_{iK} + \frac{1+\alpha}{2}b_{iK} \right]\right), \alpha\in[0,1],
    \end{equation}
    then it is easy to see that $\mu_L(\gT_{\alpha})={\mu_L(\gS)}/2$ and that both $\gT_{\alpha}$ and $\gS\setminus\gT_{\alpha}$ are hyperrectangle collections.\footnote{Intuitively, $\gT_\alpha$ splits $\gS$ along the $K$-th dimension into two sets: one set that contains an interval of length $(b_{iK}-a_{iK})/2$ that lies strictly inside the interval $[a_{iK}, b_{iK}]$ and one set that contains two disconnected intervals, one to the left of the aforementioned middle interval and of length $\alpha(b_{iK}-a_{iK})/2$ and one to the right of the aforementioned interval of length $\left(1-\alpha\right)\,(b_{iK}-a_{iK})/2$.}

    For $\alpha_0\in(0,1)$, dominated convergence theorem implies that $\lim_{\alpha\to\alpha_0}\int_{\gT_{\alpha}} w(x) \text{d} \mu_L(x) = \lim_{\alpha\to\alpha_0}\int_{\gS} \mathbbm{1}\{x\in\gT_{\alpha}\} w(x) \text{d} \mu_L(x) = \int_{\gS} \mathbbm{1}\{x\in\gT_{\alpha_0}\} w(x) \text{d} \mu_L(x) = \int_{\gT_{\alpha_0}} w(x) \text{d} \mu_L(x)$, so
    the mapping $\psi: [0,1] \mapsto \R,\quad \alpha \mapsto \int_{\gT_{\alpha}} w(x) \text{d} \mu_L(x)$
    is continuous and satisfies $\psi(0)+\psi(1) = \int_{\gS} w(x) \text{d} \mu_L(x)$,
    because $\gT_0\cup\gT_1=\gS$ and $\mu_L(\gT_0\cap\gT_1)=0$, so there must exists some $\alpha$ such that $\psi(\alpha)=\frac{\psi(0)+\psi(1)}{2} = \frac{1}{2} \int_{\gS} w(x) \text{d} \mu_L(x)$.
    Hence we can choose $\gS_1=\gT_{\alpha}$ and $\gS_2=\gS\setminus\gT_{\alpha}$, concluding the proof.

\subsubsection{Proof of Lemma \ref{lemma:nuisance-radius-1}}

We have
\begin{equation}
    \notag
    \normx{m_{\lambda}(X)-\hat{m}(X)}_{P_X,r} =\beta\left\Vert \frac{\hat{w}(X)\Delta(\lambda,X)}{\hat{g}(1,X)}\right\Vert _{P_X,r} \leq c^{-1}\beta \Vert \hat{w}(X)\Vert _{P_X,r}
\end{equation}
and
\begin{equation}
    \notag
    \begin{aligned}
        \normx{g_{\lambda}(1,X)-\hat{g}(1,X)}_{P_X,r} 
        &\leq  \left\Vert \frac{\hat{m}(X)-m_{\lambda}(X)}{m_{\lambda}(X)}\hat{g}(1,X)\right\Vert _{P_X,r} + \alpha \left\Vert \frac{\hat{m}(X)}{m_{\lambda}(X)}\hat{w}(X)\right\Vert _{P_X,r} \\
        &\leq 2(c^{-1}\beta + \alpha)\Vert \hat{w}(X)\Vert _{P_X,r}.
    \end{aligned}
\end{equation}

\subsubsection{Proof of Lemma \ref{lemma:ate-case1-mixture-equal}}

By definition, we have 
\begin{equation}
    \notag
    q_{\lambda}(x,d,y) =  m_{\lambda}(x)^d (1-m_{\lambda}(x))^{1-d} g_{\lambda}(d,x)^y (1-g_{\lambda}(d,x))^{1-y}
\end{equation}
and
\begin{equation}
    \notag
    \hat{p}(x,d,y) = \hat{m}(x)^d(1-\hat{m}(x))^{1-d}\hat{g}(d,x)^y(1-\hat{g}(d,x))^{1-y}.
\end{equation}
The mixed joint density $q$ is then given by
\begin{equation}
    \notag
    \begin{aligned}
        q(x,d,y) &= \int q_{\lambda}(x,d,y) \text{d} \pi(\lambda) = \int m_{\lambda}(x)^d (1-m_{\lambda}(x))^{1-d} g_{\lambda}(d,x)^y (1-g_{\lambda}(d,x))^{1-y} \text{d} \pi(\lambda)
    \end{aligned}
\end{equation}
When $d=1$, we have
\begin{equation}
    \notag
    \begin{aligned}
        q(x,1,y) 
        &= \left\{
            \begin{aligned}
                \int m_{\lambda}(x) g_{\lambda}(1,x) \text{d} \pi(\lambda) \quad &\text{if } y=1 \\
                \int m_{\lambda}(x) \left(1-g_{\lambda}(1,x)\right) \text{d} \pi(\lambda) \quad &\text{if } y=0.
            \end{aligned}
        \right.
    \end{aligned}
\end{equation}
By \eqref{eq:lambda-avg-1}, we know that $\int m_{\lambda}(x) g_{\lambda}(1,x) \text{d} \pi(\lambda) = \hat{m}(x)\hat{g}(1,x)=\hat{p}(x,1,1)$
and $\int m_{\lambda}(x) \left(1-g_{\lambda}(1,x)\right) \text{d} \pi(\lambda) = \hat{m}(x)-\hat{m}(x)\hat{g}(1,x)=\hat{p}(x,1,0),$
thus $q(x,1,y)=\hat{p}(x,1,y), y\in\{0,1\}$. 

When $d=0$, recall that $\hat{g}(0,x)=g_{\lambda}(0,x)$ by our construction, so we have
\begin{equation}
    \notag
    \begin{aligned}
        q(x,0,y) &= \int (1-m_{\lambda}(x))\hat{g}(0,x)^y(1-\hat{g}(0,x))^{1-y} \text{d} \pi(\lambda)  \\
        &= (1-\hat{m}(x))\hat{g}(0,x)^y(1-\hat{g}(0,x))^{1-y} = \hat{p}(x,0,y).
    \end{aligned}
\end{equation}
where we again use \eqref{eq:lambda-avg-1-m} in the second equation. Hence $\hat{p}=q$ as desired.

\subsubsection{Proof of Lemma \ref{lemma:ate-case1-hellinger-bound}}

We prove this lemma by applying Lemma \ref{semi-param-thm} to the partition $\gX_j = \left(B_{2j-1}\cup B_{2j}\right)\times\{0,1\}\times\{0,1\},\quad j=1,2,\cdots,M/2$
    of $[0,1]^K\times\{0,1\}\times\{0,1\}$, $p=\hat{p}$ and $q_{\lambda}$ as constructed above, and $\mu$ being the uniform distribution over $[0,1]^K\times\{0,1\}\times\{0,1\}$. Recall that $B_j$'s are chosen to satisfy
    $\mu_L(B_j) = 1/M$
    where $\mu_L$ is the Lebesgue measure, so that
    \begin{equation}
        \label{eq:pj-value}
        p_j := \hat{P}(\gX_j) = Q_{\lambda} (\gX_j) = \mu_L(B_{2j-1})+\mu_L(B_{2j}) = 2/M
    \end{equation}
    since their marginal distribution $P_X$ is the uniform distribution. Also, since for any $x\in\gX_j$ we have $\Delta(\lambda,x)=\lambda_j(\mathbbm{1}\{x\in B_{2j-1}\}-\mathbbm{1}\{x\in B_{2j}\})$, the distribution $Q_{\lambda}$ restricted to $\gX_j$ only depends on $\lambda_j$. 
    It follows from \eqref{eq:pj-value} that
    \begin{equation}
        \notag
        \begin{aligned}
            b &= \frac{M}{2} \max _j \sup _\lambda \int_{\mathcal{X}_j} \frac{\left(q_\lambda-\hat{p}\right)^2}{\hat{p}} d \mu \leq \max_j \frac{M}{2} p_j\cdot \sup_{(x,d,y)\in\gX_j} \frac{(\hat{p}(x,d,y)-q_{\lambda}(x,d,y))^2}{\hat{p}(x,d,y)} \leq \frac{4}{c^2},
        \end{aligned}
    \end{equation}
    where the last step holds since $\hat{p}(x,1,y) \geq p_X(x)\cdot\min\left\{\hat{m}(x),1-\hat{m}(x)\right\}\cdot\min\left\{\hat{g}(1,x),1-\hat{g}(1,x)\right\} \geq c^2$
    by Assumption~\ref{assm:one-side-overlap}, which implies that
    ${(\hat{p}(x,1,y)-q_{\lambda}(x,1,y))^2}/{\hat{p}(x,1,y)} \leq 4c^{-2}$
    and for all $(x,0,y)\in\mathrm{supp}(\hat{P})$,
    \begin{equation}
        \notag
        \begin{aligned}
            \frac{(\hat{p}(x,0,y)-q_{\lambda}(x,0,y))^2}{\hat{p}(x,0,y)} &\leq \frac{(m_{\lambda}(x)-\hat{m}(x))^2\hat{g}(0,x)^{2y}(1-\hat{g}(0,x))^{2(1-y)}}{(1-\hat{m}(x))\hat{g}(0,x)^{y}(1-\hat{g}(0,x))^{1-y}} \leq \frac{4}{c}.
        \end{aligned}
    \end{equation}
    Hence we have $C n^2 \left(\max_j p_j\right)b^2 \leq 32Cn^2/(c^4 M) \leq \delta$.
    Finally, we have $n\max\{1,b\}\max_j p_j \leq 4nc^{-2}M^{-1} \leq 4c^{-2}=A$ by our choice of $M$, so all conditions of Lemma~\ref{semi-param-thm} hold.
    By Lemma \ref{semi-param-thm}, we can conclude that $H^2(\hat{P},Q) \leq \delta$.

\subsubsection{Proof of Lemma \ref{lemma:ate-case1-bounds-collect}}

Our assumption that $e_n' \geq f_n$ implies that $\alpha\geq\beta$.
Since $e_n', f_n = o(1) (n\to +\infty)$, for sufficiently large $n$ we must have 
\begin{equation}
    \label{eq:ate-1-beta-small}
    \max\{\alpha, \beta\}\leq \frac{1}{4}c^2\big(1+\Vert w(X)\Vert _{P_X,\infty}\big)^{-4}\min\left\{1,\E\left[\frac{w(X)\hat{w}(X)^2}{\hat{g}(1,X)}\right]\right\},
\end{equation}
where $c$ is the constant introduced in Assumption \ref{assm:one-side-overlap}. In the remaining part of the proof we will assume that \eqref{eq:ate-1-beta-small} holds.  

First, by Lemma \ref{lemma:nuisance-radius-1} and our choice of $\alpha$ and $\beta$ it is easy to see that
\begin{equation}
    \notag
    \begin{aligned}
        &\normx{\hat{m}(X)-m_{\lambda}(X)}_{P_X,2} \leq c^{-1} \beta \Vert \hat{w}(X)\Vert _{P_X,2}  \leq \sqrt{f_n} \\
        &\normx{\hat{g}(1,X)-g_{\lambda}(1,X)}_{P_X,2} \leq (\alpha+c^{-1}\beta)\Vert \hat{w}(X)\Vert _{P_X,2} \leq \sqrt{e_n'}. 
    \end{aligned}
\end{equation}
Note that the second inequality above makes use of our assumption that $e_n'\geq f_n$. Again applying Lemma \ref{lemma:nuisance-radius-1} with $r=\infty$, we have 
\begin{equation}
    \notag
    \normx{\hat{g}(1,X)-g_{\lambda}(1,X)}_{P_X,\infty} \leq (\alpha+c^{-1}\beta)\Vert \hat{w}(X)\Vert _{P_X,\infty} \leq c/2,
\end{equation}
which implies that $0\leq g_{\lambda}\leq 1$. Similarly we have $0\leq m_{\lambda}\leq 1$, so $(m_{\lambda}, g_{\lambda})\in\gF_{e_n,e_n',f_n}$.

It remains to show that \eqref{eq:list-cond-1.3} holds. To see this, note that for fixed $\lambda\in\{0,1\}^{M/2}$ we have
\begin{subequations}
    \label{eq:ate1-taylor}
    \begin{align}
        &\quad \E\left[ w(X) g_{\lambda}(1,X) \right]\nonumber \\
        &= \E\left[ w(X) \frac{\hat{g}(1,X) + \alpha\hat{w}(X)\Delta(\lambda,X)}{1 -  \frac{\beta}{\hat{g}(1,X)}\hat{w}(X)\Delta(\lambda,X)} \right] \label{eq:ate1-taylor-1} \\
        &= \E\left[ w(X) \left(\hat{g}(1,X) + \alpha\hat{w}(X)\Delta(\lambda,X)\right)\sum_{k=0}^{+\infty} \left(\frac{\beta}{\hat{g}(1,X)}\hat{w}(X)\Delta(\lambda,X)\right)^k \right] \label{eq:ate1-taylor-2} \\
        &= \E \left[ w(X)\left( \hat{g}(1,X) + (\alpha+\beta)\hat{w}(X)\Delta(\lambda,X) + \frac{\alpha\beta+\beta^2}{\hat{g}(1,X)}\hat{w}(X)^2 \right) + \frac{\alpha\beta^2}{\hat{g}(1,X)^2}w(X)\hat{w}(X)^3\Delta(\lambda,X) \right] \nonumber \\
        &\quad + \E\left[ w(X) \left(\hat{g}(1,X) + \alpha\hat{w}(X)\Delta(\lambda,X)\right)\sum_{k=3}^{+\infty} \left(\frac{\beta}{\hat{g}(1,X)}\hat{w}(X)\Delta(\lambda,X)\right)^k \right] \label{eq:ate1-taylor-3} \\
        &\geq \E\left[w(X)\hat{g}(1,X) \right] + \alpha\beta \E\left[\frac{w(X)\hat{w}(X)^2}{\hat{g}(1,X)}\right] - c^{-2}\alpha\beta^2\Vert w\Vert _{P_X,\infty}^4-2c\Vert w\Vert _{P_X,\infty}\sum_{k=3}^{+\infty}\left(c^{-1}\beta\Vert w\Vert _{P_X,\infty}\right)^k \label{eq:ate1-taylor-4} \\
        &\geq \E\left[w(X)\hat{g}(1,X) \right] + \alpha\beta \E\left[\frac{w(X)\hat{w}(X)^2}{\hat{g}(1,X)}\right] - C_0(\alpha\beta^2+\beta^3), \label{eq:ate1-taylor-5}
    \end{align}
\end{subequations}
where \eqref{eq:ate1-taylor-1} follows from our construction in \eqref{eq:ate-construction-1}, \eqref{eq:ate1-taylor-2} uses a Taylor expansion which is valid since \eqref{eq:ate-1-beta-small} implies that $\left|\beta\hat{w}(X)\Delta(\lambda,X)/{\hat{g}(1,X)}\right| \leq c^{-1}\Vert w\Vert _{P_X,\infty}\beta \leq \frac{1}{2}$, \eqref{eq:ate1-taylor-3} follows from a direct expansion of \eqref{eq:ate1-taylor-2} up to the second-order term, \eqref{eq:ate1-taylor-4} is deduced by noticing that $\E_{X}[w(X)\hat{w}(X)\Delta(\lambda,X)]=0$ (by Proposition~\ref{prop:delta-prop}) and ${w(X)\hat{w}(X)^2}/{\hat{g}(1,X)}\geq 0$ and $\hat{g}(1,X)\geq c$ and using the upper bound on $\alpha$ by \eqref{eq:ate-1-beta-small}. Finally, \eqref{eq:ate1-taylor-5} holds for $C_0=4c^{-2}\Vert w\Vert _{P_X,\infty}^4$, invoking also the identity $\sum_{k=3}^{\infty} t^k = t^3 / (1-t)$ for $t=c^{-1}\beta\|w\|_{P_X,\infty}\leq 1/2$. Here, it is important to note that our construction in \eqref{eq:ate-construction-1} exactly ensures that the first-order terms (in $\alpha$ and $\beta$) cancel out. Finally, \eqref{eq:ate-1-beta-small} and $\alpha\geq\beta$ together imply that $C_0(\alpha\beta^2+\beta^3) \leq 2C_0\alpha\beta^2\leq \frac{1}{2}\E\left[{w(X)\hat{w}(X)^2}/{\hat{g}(1,X)}\right]\alpha\beta$, so \eqref{eq:list-cond-1.3} immediately follows from \eqref{eq:ate1-taylor}, concluding the proof.

\subsection{Completing the proof of the $f_n>e_n'$ case}
\label{appsubsec:case2-complete}

Parallel to Proposition \ref{prop:lambda-avg-1} and Lemma \ref{lemma:nuisance-radius-1}, we first prove some basic properties of our construction.

\begin{proposition}
\label{prop:lambda-avg-2}
    We have 
    \begin{equation}
        \notag
        \begin{aligned}
            \E_{\lambda}\left[m_{\lambda}(x)g_{\lambda}(1,x) \right] = \hat{m}(x)\hat{g}(1,x) \quad \text{and} \quad
            \E_{\lambda}\left[m_{\lambda}(x)\right] = \hat{m}(x)
        \end{aligned}
    \end{equation}
\end{proposition}

\begin{proof}
    By Proposition \ref{prop:delta-prop}, we have
    \begin{equation}
        \notag
        \begin{aligned}
            & \E_{\lambda}\left[m_{\lambda}(x)g_{\lambda}(1,x) \right] = \hat{m}(x)\hat{g}(1,x) + \alpha\hat{m}(x)\hat{g}(1,x)^2\hat{w}(x)\E_{\lambda}\Delta(\lambda,x) = \hat{m}(x)\hat{g}(1,x) \\
            & 
            \begin{aligned}
                \E_{\lambda}\left[m_{\lambda}(x)\right] &= \E_{\lambda}\Big[\left(\hat{m}(x) + \alpha\hat{m}(x)\hat{g}(1,x)\hat{w}(x)\Delta(\lambda,x)\right)\Big(1+\frac{\beta}{\hat{g}(1,x)}\hat{w}(x)\Delta(\lambda,x)-\alpha\beta\hat{w}(x)^2\Big) \Big] \\
            &= \hat{m}(x) + \left[\alpha(1-\alpha\beta\hat{w}(x)^2)\hat{m}(x)\hat{g}(1,x)\hat{w}(x)+\beta\frac{\hat{m}(x)}{\hat{g}(1,x)}\hat{w}(x)\right]\E_{\lambda}\Delta(\lambda,x) \\
            &\quad - \alpha\beta\hat{m}(x)\left(1-\E_{\lambda}\Delta(\lambda,x)^2\right)\hat{w}(x)^2 = \hat{m}(x).
            \end{aligned}
        \end{aligned}
    \end{equation}
\end{proof}

\begin{lemma}
    \label{lemma:nuisance-radius-2}
    Assuming that $\alpha\leq \max\{1,\Vert w\Vert_{P_X,\infty}\}^{-1}$ and $\beta\leq c\max\{1,\Vert w\Vert_{P_X,\infty}\}^{-2}/4$ where $c$ is a constant introduced in Assumption \ref{asmp:estimator-bounded}, then the following holds for all $0<r\leq+\infty$:
    \begin{equation}
        \notag
        \begin{aligned}
            \normx{g_{\lambda}(1,X)-\hat{g}(1,X)}_{P_X,r} &\leq 2\beta\Vert \hat{w}(X)\Vert _{P_X,r} \\
            \normx{m_{\lambda}(X)-\hat{m}(X)}_{P_X,r} &\leq 2 (\alpha+c^{-1}\beta)\Vert \hat{w}(X)\Vert _{P_X,r}.
        \end{aligned}
    \end{equation}
\end{lemma}

\begin{proof}
    From our assumptions on $\alpha$ and $\beta$, one can see that $\big| \beta(\hat{w}(x)\Delta(\lambda,x)-\alpha\beta\hat{w}(x)^2)/{\hat{g}(1,x)} \big| \leq c^{-1}\beta\|w\|_{P_X,\infty}+\beta\|w\|_{P_X,\infty}^2 \leq{1}/{2}$. Thus it follows that
    \begin{equation}
        \notag
        \begin{aligned}
            \normx{g_{\lambda}(1,X)-\hat{g}(1,X)}_{P_X,r} &\leq 2\normx{ \hat{g}(1,X)\bigg( \frac{\beta}{\hat{g}(1,X)}\hat{w}(X)\Delta(\lambda,X)-\alpha\beta\hat{w}(X)^2\bigg)}_{P_X,r} \\
            &\leq (\beta+\alpha\beta\|w\|_{P_X,\infty})\|w(X)\|_{P_X,r} \leq 2\|w(X)\|_{P_X,r}
        \end{aligned}
    \end{equation}
    and
    \begin{equation}
        \notag
        \begin{aligned}
            \normx{m_{\lambda}(X)-\hat{m}(X)}_{P_X,r} &\leq 2c^{-1}\beta\|w\|_{P_X,\infty}+\alpha\|w\|_{P_X,\infty}+2c^{-1}\alpha\beta\|w\|_{P_X,\infty}^2 \\
            &\leq 2 (\alpha+c^{-1}\beta)\Vert \hat{w}(X)\Vert _{P_X,r}.
        \end{aligned}
    \end{equation}
\end{proof}

Note that a key difference between Lemma \ref{lemma:nuisance-radius-2} and Lemma \ref{lemma:nuisance-radius-1} is that in the former lemma, the deviations of $g_{\lambda}$ and $m_{\lambda}$ are $\gO(\beta)$ and $\gO(\alpha+\beta)$ respectively, while the converse is true in the latter one. This difference is intentional, since here we assume that $f_n>e_n'$.

Let $Q_{\lambda}$ be the joint distribution of $(X,D,Y)$ induced by $g_{\lambda}$ and $m_{\lambda}$ and $\mu$ be the uniform distribution on $[0,1]^K\times\{0,1\}\times\{0,1\}$. Define $q_{\lambda} = {\text{d} Q_{\lambda}}/{\text{d} \mu}$.
Similarly, let $\hat{P}$ be the joint distribution of $(X,D,Y)$ induced by $\hat{g}$ and $\hat{m}$, and $\hat{p}={\text{d} \hat{P}}/{\text{d}\mu}$. Using exactly the same arguments as we did in Lemma \ref{lemma:ate-case1-mixture-equal} and \ref{lemma:ate-case1-hellinger-bound}, one can prove the following lemmas.

\begin{lemma}
    \label{lemma:ate-case2-mixture-equal}
    Let $Q = \int Q_{\lambda} \text{d} \pi(\lambda)$ and $q = {\text{d} Q}/{\text{d} \mu} = \int q_{\lambda} \text{d} \pi(\lambda)$, then $\hat{p}=q$.
\end{lemma}

\begin{lemma}
\label{lemma:ate-case2-hellinger-bound}
    For any $\delta>0$, as long as $M \geq \max\{n,32Cn^2/(c^4\delta)\}$ where $c$ is the constant introduced in Assumption~\ref{assm:one-side-overlap} and $C$ is the constant implied by Lemma~\ref{semi-param-thm} for $A=4c^{-2}$, we have $H^2\left(\hat{P}^{\otimes n},\int Q_{\lambda}^{\otimes n}\text{d} \pi(\lambda)\right) \leq \delta.$
\end{lemma}

Finally, we prove the analogue of Lemma \ref{lemma:ate-case1-bounds-collect} for the different construction that we are now considering.

\begin{lemma}
\label{lemma:ate-case2-bounds-collect}
    Let $\alpha = \sqrt{f_n}/\big(4\Vert \hat{w}(X)\Vert _{P_X,2}\big),\quad \beta = c\sqrt{e_n'}/\big(4\Vert \hat{w}(X)\Vert _{P_X,2}\big),$
    then for sufficiently large $n$, we have $(m_{\lambda},g_{\lambda})\in\gF_{e_n,e_n',f_n}$, and $\forall\lambda\in\{0,1\}^{M/2}$:
    \begin{equation}
    \label{eq:list-cond-2.3}
        \E_X\left[w(X) g_{\lambda}(1,X)\right] \geq \E\left[w(X) \hat{g}(1,X) \right] + \alpha\beta/2\cdot\E_X \left[\hat{g}(1,X)w(X)\hat{w}(X)^2\right] 
    \end{equation}
\end{lemma}

\begin{proof}
    Since $e_n', f_n = o(1) (n\to +\infty)$, for sufficiently large $n$ we must have 
    \begin{equation}
        \label{eq:ate-2-beta-small}
        \max\{\alpha,\beta\} < c^2/4\cdot(1+\Vert w\Vert _{P_X,\infty})^{-4}\min\left\{1,\E_X \left[\hat{g}(1,X)w(X)\hat{w}(X)^2\right]\right\}, 
    \end{equation}
    where $c$ is the constant introduced in Assumption \ref{assm:one-side-overlap}. First, by Lemma \ref{lemma:nuisance-radius-2} our choice of $\alpha$ and $\beta$ it is easy to see that
    \begin{equation}
        \notag
        \begin{aligned}
            &\normx{\hat{m}(X)-m_{\lambda}(X)}_{P_X,2} \leq 2(\alpha+c^{-1}\beta) \Vert \hat{w}(X)\Vert _{P_X,2} \leq \sqrt{f_n} \\
            &\normx{\hat{g}(1,X)-g_{\lambda}(1,X)}_{P_X,2}  \leq 2\beta \Vert \hat{w}(X)\Vert _{P_X,2} \leq \sqrt{e_n'}. 
        \end{aligned}
    \end{equation}
    Note that the first inequality above makes use of our assumption that $f_n> e_n'$. Applying Lemma \ref{lemma:nuisance-radius-2} with $r=\infty$, \eqref{eq:ate-2-beta-small} implies that $0\leq m_{\lambda},g_{\lambda}\leq 1$. Hence $(m_{\lambda},g_{\lambda})\in\gF_{e_n,e_n',f_n}$.
    
    It remains to show that \eqref{eq:list-cond-2.3} holds. Note that for fixed $\lambda\in\{0,1\}^{M/2}$ we have
    \begin{subequations}
        \label{eq:ate2-taylor}
        \begin{align}
            &\quad \E_{X}\left[w(X) g_{\lambda}(1,X)\right] \nonumber\\ 
            &= \E_X \left[ w(X) \frac{\hat{g}(1,X)}{1+\frac{\beta}{\hat{g}(1,X)}\hat{w}(X)\Delta(\lambda,X)-\alpha\beta\hat{w}(X)^2} \right]\nonumber \\
            &= \E_{X}\left[ w(X) \hat{g}(1,X)\left( 1 + \sum_{k=1}^{+\infty}\beta^k \left(\alpha\hat{w}(X)^2-\frac{1}{\hat{g}(1,X)}\hat{w}(X)\Delta(\lambda,X)\right)^k \right) \right]\label{eq:ate2-taylor-1} \\
            &= \E_X w(X)\hat{g}(1,X) + \alpha\beta\E_X\left[ \hat{g}(1,X)w(X)\hat{w}(X)^2 \right] -\beta\E_X\left[w(X)\hat{w}(X)\Delta(\lambda,X)\right]\nonumber \\
            &\quad + \E_{X}\left[ w(X) \hat{g}(1,X)\sum_{k=2}^{+\infty}\beta^k \left(\alpha\hat{w}(X)^2-\frac{1}{\hat{g}(1,X)}\hat{w}(X)\Delta(\lambda,X)\right)^k  \right]\label{eq:ate2-taylor-2} \\
            &\geq \E_X\left[w(X)\hat{g}(1,X)\right] + \alpha\beta\E_X \left[\hat{g}(1,X)w(X)\hat{w}(X)^2\right] - C_0\beta^3\label{eq:ate2-taylor-3},
        \end{align}
    \end{subequations}
    where \eqref{eq:ate2-taylor-1} uses Taylor expansion which holds since
    \begin{equation}
        \notag
        \left|\beta\hat{w}(X)\Big(\alpha\hat{w}(X)-\frac{1}{\hat{g}(1,X)}\Delta(\lambda,X)\Big)\right|\leq\frac{1}{4}c\cdot\left(1+\frac{1}{c}\right)\leq\frac{1}{2}
    \end{equation}
    by \eqref{eq:ate-2-beta-small}, \eqref{eq:ate2-taylor-2} follows from directly expanding \eqref{eq:ate2-taylor-1}, and \eqref{eq:ate2-taylor-3} holds with $C_0=2c^{-2}\Vert w\Vert _{P_X,\infty}^4$ where we use the fact that $\E_X\left[w(X)\hat{w}(X)\Delta(\lambda,X)\right]=0$ (by Proposition~\ref{prop:delta-prop}) and that for any $|t|\leq 1/2$, $\sum_{k=2}^{\infty}t^k \geq \sum_{k=3}^{\infty}t^k = t^3 / (1-t)$ (applied for $t:=\beta\left(\alpha\hat{w}(X)^2-{\hat{g}(1,X)}^{-1}\hat{w}(X)\Delta(\lambda,X)\right)$, which also satisfies that $t^3 \geq -\beta^3 \|w\|_{P_X,\infty}^3\hat{g}(1,X)^{-3}$). Moreover, \eqref{eq:ate-2-beta-small} and $f_n>e_n'$ together imply that $C_0\beta^3 \leq C_0 \beta^2 c \alpha \leq c \alpha\beta/2\cdot\E_X \left[\hat{g}(1,X)w(X)\hat{w}(X)^2\right]$, so \eqref{eq:list-cond-2.3} immediately follows from \eqref{eq:ate2-taylor}, concluding the proof.
\end{proof}

We are now ready to prove Theorem \ref{thm:ate} in the case when $f_n>e_n$. We choose $M \geq \max\{n,32Cn^2/(c^4\delta)\}$ and $\gP=\{\hat{P}\}\cup\left\{Q_{\lambda}:\lambda\in\{0,1\}^{M/2}\right\}$, $P=\hat{P}$, $\pi$ be the discrete uniform distribution on  $\left\{Q_{\lambda}:\lambda\in\{0,1\}^{M/2}\right\}$, $s=\alpha\beta/4\cdot\E_X \left[\hat{g}(1,X)w(X)\hat{w}(X)^2\right]$ in the context of Lemma \ref{fano-method}. Then all the listed conditions are satisfied for the \textsc{WATE} functional
\begin{equation}
    \notag
    T(P) = \theta^{\wate}(P) = \E_P\left[ w(x)\left(g(1,X)-g(0,X)\right)\right].
\end{equation}
Therefore, by Lemma \ref{fano-method}, we obtain a lower bound
\begin{equation}
    \notag
    \begin{aligned}
        &\quad \inf_{\hat{\theta}}\sup_{P\in\gP} \gQ_{P,1-\gamma}\left(\left| \hat{\theta}\left(\{(X_i,D_i,Y_i)\}_{i=1}^N\right) - \theta^{\wate}\right|^2\right) \\
        &= \Omega\left(\alpha\beta\E_X \left[\hat{g}(1,X)w(X)\hat{w}(X)^2\right]\right) = \Omega\left( \sqrt{e_n' f_n}\cdot\frac{\E_X \left[\hat{g}(1,X)w(X)\hat{w}(X)^2\right]}{\Vert \hat{w}(x)\Vert _{P,2}^2}\right)\\
        &= \Omega\left( \sqrt{e_n' f_n}\cdot\frac{\E_X \left[w(X)^3\mathbbm{1}\left(w(X)\geq\frac{1}{2}\Vert w\Vert _{P_X,\infty}\right)\right]}{\E_X \left[w(X)^2\mathbbm{1}\left(w(X)\geq\frac{1}{2}\Vert w\Vert _{P_X,\infty}\right)\right]}\right) = \Omega \left(\Vert w\Vert _{P_X,\infty}\cdot\sqrt{e_n'f_n} \right).
    \end{aligned}
\end{equation}

Combining the derivations in Section \ref{subsec:case1} and \ref{subsec:case2}, we have shown that
$\mathfrak{M}_n^{\wate}\left(\gF_{e_n,e_n',f_n}\right) = \Omega\left( e_n' f_n\Vert w\Vert _{P_X,\infty}^2 \right).$ Using similar arguments, one can establish the lower bound $\Omega\left( e_n f_n\Vert w\Vert _{P_X,\infty}^2 \right)$; details can be found in the supplementary material Section S2. Finally, the $\Omega(1/\sqrt{n})$ rate is standard and the proof is given in supplementary material Section S4.

\subsection{Proof of the lower bound $e_nf_n\Vert w\Vert _{P_X,\infty}^2$ in Theorem \ref{thm:ate}}

In this section, we illustrate how the lower bound $\Omega\left( e_n f_n \Vert w\Vert _{P_X,\infty}^2 \right)$ can be derived in a completely symmetric fashion. Parallel to the proofs in Section 4.4 and 4.5, we also consider two cases: $e_n\geq f_n$ and $e_n<f_n$.

In the first case, we define
\begin{equation}
    \label{eq:ate-construction-3}
    \begin{aligned}
        & g_{\lambda}(0,x) = \frac{1-\hat{m}(x)}{1-m_{\lambda}(x)} \left[\hat{g}(0,x) - \alpha\hat{w}(x)\Delta(\lambda,x) \right],\quad g_{\lambda}(1,x) = \hat{g}(1,x), \\
        & m_{\lambda}(x) = \hat{m}(x) + \left(1-\hat{m}(x)\right)  \frac{\beta}{\hat{g}(0,x)}\hat{w}(x)\Delta(\lambda,x).
    \end{aligned}
\end{equation}

In the second case, we define
\begin{equation}
    \label{eq:ate-construction-4}
    \begin{aligned}
        & g_{\lambda}(0,x) = \frac{\hat{g}(0,x)}{1+\frac{\beta}{\hat{g}(0,x)}\hat{w}(x)\Delta(\lambda,x)-\alpha\beta\hat{w}(x)^2} ,\quad g_{\lambda}(1,x) = \hat{g}(1,x),  \\
        & m_{\lambda}(x) = 1- \frac{\hat{g}(0,x)}{g_{\lambda}(0,x)}(1-\hat{m}(x))\left(1 - \alpha\hat{g}(0,x)\hat{w}(x)\Delta(\lambda,x)\right).
    \end{aligned}
\end{equation}

Then we have the following result.

\begin{lemma}
\label{lemma:rn'-collect}
    Let $Q_{\lambda}$ be the joint distribution of $(X,D,Y)$ induced by $g_{\lambda}$ and $m_{\lambda}$ and $\mu$ be the uniform distribution on $[0,1]^K\times\{0,1\}\times\{0,1\}$. Define $q_{\lambda} = {\text{d} Q_{\lambda}}/{\text{d} \mu}$. Then $\int q_{\lambda}\text{d}\pi(\lambda)=\hat{p}$.
    Moreover, there exists constants $c_{\alpha}, c_{\beta} > 0$, such that by choosing 
    \begin{equation}
        \notag
        (\alpha,\beta) = \left\{
        \begin{aligned}
            \left(c_{\alpha}\frac{\sqrt{e_n}}{\Vert \hat{w}(X)\Vert _{P_X,2}},c_{\beta}\frac{\sqrt{f_n}}{\Vert \hat{w}(X)\Vert _{P_X,2}}\right) &\quad\text{for the first case;} \\
            \left(c_{\alpha}\frac{\sqrt{f_n}}{\Vert \hat{w}(X)\Vert _{P_X,2}},c_{\beta}\frac{\sqrt{e_n}}{\Vert \hat{w}(X)\Vert _{P_X,2}}\right) &\quad\text{for the second case,} 
        \end{aligned}
        \right.
    \end{equation}
    the following inequalities hold for sufficiently large $n$:
    \begin{subequations}
    \label{eq:list-cond-11}
    \begin{align}
        &\normx{\hat{m}(X)-m_{\lambda}(X)}_{P_X,2}^2 \leq f_n, \label{eq:list-cond-11.1} \\
        &\normx{\hat{g}(0,X)-g_{\lambda}(0,X)}_{P_X,2}^2 \leq e_n, \label{eq:list-cond-11.2} \\
        &\E_X\left[w(X) g_{\lambda}(0,X)\right] \leq \E\left[w(X) \hat{g}(0,X) \right] - \Omega\left( \sqrt{e_nf_n}\Vert w\Vert _{P_X,\infty}\right). \label{eq:list-cond-11.3}
    \end{align}
\end{subequations}
\end{lemma}

The proof of Lemma \ref{lemma:rn'-collect} follows the exactly same route as the proofs in Section 4.4 and 4.5, so we do not repeat it here. Finally, we can directly apply Theorem 4 to obtain the lower bound $\Omega\left( e_n f_n\Vert w\Vert _{P_X,\infty}^2 \right)$.

\section{Proof of Theorem \ref{thm:atte}}
\label{subsec:proof-atte}

In this section, we give the detailed proof of our main result, Theorem \ref{thm:atte}, for the lower bound of estimating \textsc{ATT}. The idea of the proof is similar to that of Theorem 1, but additional effort needs to be made to guarantee that the separation condition \eqref{fano:separation-condition} holds.

Let $P_X$ be the uniform distribution on $\mathrm{supp}(X)=[0,1]^K$, and $[0,1]^K$ be partitioned into $M$ cubes $B_1, B_2,\cdots, B_M$, each with volume ${1}/{M}$. Let $\lambda_i, i=1,2,\cdots, {M}/{2}$ be i.i.d. variables taking values $+1$ and $-1$ both with probability $0.5$. 

Define
\begin{equation}
    \notag
    \theta_{\ml}^{\att} = \left(\E_X\left[\hat{m}(X)\right]\right)^{-1}\E_X\left[\hat{m}(X)\left(\hat{g}(1,X)-\hat{g}(0,X)\right)\right].
\end{equation}
We first prove the following lemma:

\begin{lemma}
\label{lemma:choose-u}
    There exist constants $C_u, c_u > 0$ that only depend on $\hat{m}$ and $\hat{g}$, such that for all sufficiently large integer $M$, there exists a function $u:[0,1]^K\to\mathbb{R}_{\geq 0}$ satisfying $\normx{u}_{\infty}\leq C_u$ and a partition $[0,1]^K = \cup_{j=1}^M B_j$ into Lebesgue-measurable sets $B_j$ each with measure ${1}/{M}$, such that
    \begin{equation}
        \label{eq:choose-u-1}
        \E_X\left[ u(X)\left(\hat{g}(1,X)-\hat{g}(0,X)-\theta_{\ml}^{\att}\right)\Delta(\lambda,X) \right] = 0,\quad \forall \lambda\in\{-1,+1\}^{M/2}
    \end{equation}
    and
    \begin{equation}
        \label{eq:choose-u-2}
        \E_X\left[ \frac{u(X)}{\hat{m}(X)\left(1-\hat{m}(X)\right)} \right] \geq c_u,
    \end{equation}
    where we recall that $\Delta(\lambda,x) := \sum_{j=1}^{M/2}\lambda_j\left(\mathbbm{1}\left\{x\in B_{2j-1}\right\}-\mathbbm{1}\left\{x\in B_{2j}\right\}\right).$
\end{lemma}

\begin{proof}
    Let $\alpha = \mathbb{P}\left[  \hat{g}(1,X)-\hat{g}(0,X)-\theta_{\ml}^{\att} = 0\right]$. If $\alpha=1$, then we can simply choose $u=1$ and $c_u=1$. Thus we can assume that $\alpha<1$. In this case either
    \begin{equation}
        \notag
        \mathbb{P}\left[  \hat{g}(1,X)-\hat{g}(0,X)-\theta_{\ml}^{\att} > 0\right] \geq \frac{1-\alpha}{2}
    \end{equation}
    or
    \begin{equation}
        \notag
        \mathbb{P}\left[  \hat{g}(1,X)-\hat{g}(0,X)-\theta_{\ml}^{\att} < 0\right] \geq \frac{1-\alpha}{2}.
    \end{equation}
    We proceed by assuming that the former holds; the case when the latter holds can be handled in exactly the same way.

    Define the event $\gE_{\delta} = \left\{\hat{g}(1,X)-\hat{g}(0,X)-\theta_{\ml}^{\att} > \delta\right\}$, then $\lim_{\delta\to 0} \mathbb{P}\left[\gE_{\delta}\right] \geq {(1-\alpha)}/{2}$,
    so there exists $\delta_0>0$ such that $\mathbb{P}\left[\gE_{\delta_0}\right] \geq {(1-\alpha)}/{3}$.
    
    Let $M_{\alpha} = 2\floor{(1-\alpha)M/6}$ and let $B_j, 1\leq j\leq M$ be chosen in a way such that $B_j, 1\leq j\leq M_{\alpha}$ are (disjoint) measurable subsets of $\gE_{\delta_0}$ with measure ${1}/{M}$; the remaining $B_j$'s can be chosen arbitrarily. Then we define
    \begin{equation}
        \notag
        u(x) = \left\{
        \begin{aligned}
            & 1 & x\in B_{2j-1}, 1\leq j\leq M_{\alpha}/2 \\
            & \frac{\E_X\left[ \left(\hat{g}(1,X)-\hat{g}(0,X)-\theta_{\ml}^{\att}\right)\mathbbm{1}\left\{X\in B_{2j-1}\right\} \right]}{\E_X\left[ \left(\hat{g}(1,X)-\hat{g}(0,X)-\theta_{\ml}^{\att}\right)\mathbbm{1}\left\{X\in B_{2j}\right\} \right]}  & x\in B_{2j}, 1\leq j\leq M_{\alpha}/2 \\
            &0  & \text{otherwise}.
        \end{aligned}
        \right.
    \end{equation}
    Specifically, $u(x)$ is constant in each $B_j$. Moreover, note that the denominator in the second case is bounded away from zero, since these regions are subsets of $\gE_{\delta_0}$.
    First, it is easy to see that this choice of $u$ guarantees that
    \begin{equation}
        \notag
        \E_X\left[ u(X) \left(\hat{g}(1,X)-\hat{g}(0,X)-\theta_{\ml}^{\att}\right)\left(\mathbbm{1}\left\{X\in B_{2j-1}\right\}-\mathbbm{1}\left\{X\in B_{2j}\right\}\right)\right] = 0
    \end{equation}
    for all $j$, so that \eqref{eq:choose-u-1} holds.
    
    Second, let $C_u = \delta_0^{-1}\left( 2+\absx{\theta_{\ml}^{\att}}\right)$.
    Our choice of $B_j$ implies that for $1\leq j\leq M_{\alpha}/2$, we have
    \begin{align*}
        \E_X\left[ \left(\hat{g}(1,X)-\hat{g}(0,X)-\theta_{\ml}^{\att}\right)\mathbbm{1}\left\{X\in B_{2j}\right\} \right] \geq~& \delta_0\cdot\mathbb{P}\left[ X\in B_{2j}\right] = {\delta_0}/{M}
    \end{align*}
    and
    \begin{equation}
        \notag
        \begin{aligned}
            \E_X\left[ \left(\hat{g}(1,X)-\hat{g}(0,X)-\theta_{\ml}^{\att}\right)\mathbbm{1}\left\{X\in B_{2j-1}\right\} \right] \leq~& \big(2\sup_{d,x}\hat{g}(d,x) + \absx{\theta_{\ml}^{\att}}\big)P[X\in B_{2j-1}]\\
            \leq~& \big(2 + \absx{\theta_{\ml}^{\att}}\big)/{M}.
        \end{aligned}
    \end{equation}
    As a consequence, we have
    \begin{equation}
        \notag
        u(x) \leq C_u,\quad\forall x\in[0,1]^K.
    \end{equation}
    Finally, since $\mathbb{P}\left[ u(X)=1 \right] = {M_{\alpha}}/{2M}$ and $u(x)\geq 0$ for all $x$, we can deduce that
    \begin{equation}
        \notag
        \E_X\left[ \frac{u(X)}{\hat{m}(X)\left(1-\hat{m}(X)\right)} \right] \geq \frac{M_{\alpha}}{2M} \geq 0.1(1-\alpha).
    \end{equation}
    Hence, the $u(x)$ that we choose satisfies all the required conditions, concluding the proof.
\end{proof}

Returning to our proof of Theorem 2, let $u(x)$ and $\Delta(\lambda,x)$ be the function chosen in Lemma \ref{lemma:choose-u} and let
\begin{equation}
    \label{eq:atte-def-auxiliary}
    \begin{aligned}
        & v(x) = \frac{1-\hat{m}(x)}{\hat{m}(x)}.
    \end{aligned}
\end{equation}
We define
\begin{equation}
    \label{eq:atte-construction}
    \begin{aligned}
        & g_{\lambda}(0,x) = \hat{g}(0,x) + \alpha\frac{v(x)}{1-m_{\lambda}(x)}\Delta(\lambda,x), \quad g_{\lambda}(1,x) = \hat{g}(1,x), \\
        & m_{\lambda}(x) = \hat{m}(x)-\beta u(x)\Delta(\lambda,x).
    \end{aligned}
\end{equation}
where $\alpha,\beta$ are constants that will be specified later. Then one can easily derive the following results:

\begin{proposition}
    \label{prop:lambda-avg-atte}
    We have
    \begin{subequations}
        \notag
        \begin{align}
            \E_{\lambda}\left[m_{\lambda}(x)\right] =\hat{m}(x) \quad \text{and} \quad
            \E_{\lambda}\left[\left(1-m_{\lambda}(x)\right)g_{\lambda}(0,x)\right] &= \hat{g}(0,x)\left(1-\hat{m}(x)\right).
        \end{align}
    \end{subequations}
\end{proposition}

\begin{proof}
    By Proposition 1, we have
    \begin{equation}
        \notag
        \begin{aligned}
            \E_{\lambda}\left[m_{\lambda}(x)\right] &= \hat{m}(x)-\beta u(x)\E_{\lambda}\left[\Delta(\lambda,x)\right]=\hat{m}(x) \\
            \E_{\lambda}\left[\left(1-m_{\lambda}(x)\right)g_{\lambda}(0,x)\right] &= \hat{g}(0,x)\E_{\lambda}\left[1-m_{\lambda}(X)\right] + \alpha v(x)\E_{\lambda}\left[\Delta(\lambda,x)\right] = \hat{g}(0,x)\left(1-\hat{m}(x)\right).
        \end{aligned}
    \end{equation}
\end{proof}

As in Section 4, we can bound the $L^2$ distance between $g_{\lambda}, m_{\lambda}$ and $\hat{g},\hat{m}$ respectively.

\begin{lemma}
\label{lemma:nuisance-radius-3}
    Suppose that $\alpha\leq 1, \beta\leq\frac{1}{4}C_u^{-1}$ (where $C_u$ is defined in Lemma \ref{lemma:choose-u}), then the following holds for all $0<r\leq +\infty$:
    \begin{equation}
        \notag
        \normx{g_{\lambda}(0,X)-\hat{g}(0,X)}_{P_X,r} \leq 2c^{-1}\alpha,\quad \normx{m_{\lambda}(X)-\hat{m}(X)}_{P_X,r} \leq c^{-1} \beta.
    \end{equation}
\end{lemma}

\begin{remark}
    Due to the difference in construction, the bounds in the lemma above are in the forms of $\gO(\alpha)$ and $\gO(\beta)$ rather than $\gO(\alpha+\beta)$ and $\gO(\beta)$ that we encountered in the case of the \textsc{WATE}. This is the reason why we don't need to consider the two cases $e_n\geq f_n$ and $e_n<f_n$ separately for \textsc{ATT}.
\end{remark}

Let $Q_{\lambda}$ be the joint distribution of $(X,D,Y)$ induced by $g_{\lambda}$ and $m_{\lambda}$ and $\mu$ be the uniform distribution on $[0,1]^K\times\{0,1\}\times\{0,1\}$. Define $q_{\lambda} = {\text{d} Q_{\lambda}}/{\text{d} \mu}$.
Similarly, let $\hat{P}$ be the joint distribution of $(X,D,Y)$ induced by $\hat{g}$ and $\hat{m}$, and $\hat{p}={\text{d} \hat{P}}/{\text{d}\mu}$. Using exactly the same arguments as we did in Lemma 3 and 4, one can prove the following lemmas.

\begin{lemma}
    \label{lemma:atte-mixture-equal}
    Let $Q = \int Q_{\lambda} \text{d} \pi(\lambda)$ and $q = {\text{d} Q}/{\text{d} \mu} = \int q_{\lambda} \text{d} \pi(\lambda)$, then $\hat{p}=q$.
\end{lemma}

\begin{lemma}
\label{lemma:atte-hellinger-bound}
    For any $\delta>0$, as long as $M \geq \max\{n,32Cn^2/(c^4\delta)\}$ where $c$ is the constant introduced in Assumption 3 and $C$ is the constant implied by Lemma 1 for $A=4c^{-2}$, we have $H^2\left(\hat{P}^{\otimes n},\int Q_{\lambda}^{\otimes n}\text{d} \pi(\lambda)\right) \leq \delta.$
\end{lemma}

\begin{lemma}
\label{lemma:atte-bounds-collect}
    Let $\alpha = \frac{c}{4}\sqrt{e_n},\quad \beta = \frac{1}{4}\min\{c,c_u\} \sqrt{f_n}$, 
    then for sufficiently large $n$, we have $(m_{\lambda},g_{\lambda})\in\gF_{e_n,e_n',f_n}$ and
    \begin{equation}
    \label{eq:list-cond-3.3}
        \theta_{\lambda}^{\att}\leq \theta_{\ml}^{\att} - \frac{1}{2}c_u\alpha\beta,\quad\forall\lambda\in\{0,1\}^{M/2}, 
    \end{equation}
    where $\theta_{\lambda}^{\att} := \E_X\left[ g_{\lambda}(1,X)-g_{\lambda}(0,X) \mid D=1\right]$.
\end{lemma}

\begin{proof}
    Since $e_n,f_n=o(1) (n\to+\infty)$, we have $\alpha \leq c/4$ and $\beta\leq c_uc^3C_u^{-2}/4$ for sufficiently large $n$. In the remaining part of the proof, we assume that this inequality holds.
    
    First, by Lemma \ref{lemma:nuisance-radius-3} it is easy to see that
    \begin{equation}
        \notag
        \begin{aligned}
            &\normx{\hat{m}(X)-m_{\lambda}(X)}_{P_X,2} \leq 2c^{-1}\beta \leq \sqrt{f_n} ,\quad \normx{\hat{g}(1,X)-g_{\lambda}(1,X)}_{P_X,2} \leq 2c^{-1}\alpha \leq \sqrt{e_n} 
        \end{aligned}
    \end{equation}
    and $0\leq m_{\lambda},g_{\lambda} \leq 1$, so that $(m_{\lambda},g_{\lambda})\in\gF_{e_n,e_n',f_n}$.
    
    It remains to prove \eqref{eq:list-cond-3.3}.
    For fixed $\lambda$, we have
    \begin{equation}
        \label{eq:atte-expand}
        \begin{aligned}
            \theta_{\lambda}^{\att} &= \E_X\left[ g_{\lambda}(1,X)-g_{\lambda}(0,X) \mid D=1\right] \\
            &= \E_X\left[ \left(g_{\lambda}(1,X)-g_{\lambda}(0,X)\right)\frac{m_{\lambda}(X)}{\mathbb{P}_{\lambda}[D=1]} \right] \\
            &= \frac{\E_X\left[\left(\hat{g}(1,X)-\hat{g}(0,X)\right)m_{\lambda}(X)-\frac{\alpha v(x)m_{\lambda}(X)}{1-m_{\lambda}(x)}\Delta(\lambda,x) \right]}{\E_X\left[m_{\lambda}(X)\right]} \\
            &= \frac{\E_X\left[\left(\hat{g}(1,X)-\hat{g}(0,X)\right)\left(\hat{m}(X)-\beta u(X)\Delta(\lambda,X)\right)-\frac{\alpha v(x)m_{\lambda}(X)}{1-m_{\lambda}(x)}\Delta(\lambda,x) \right]}{\E_X\left[m_{\lambda}(X)\right]} \\
            &= \frac{\E_X\left[\left(\hat{g}(1,X)-\hat{g}(0,X)\right)\hat{m}(X)\right]-\beta\E_X\left[u(X)\left(\hat{g}(1,X)-\hat{g}(0,X)\right)\Delta(\lambda,X)\right]}{\E_X\left[\hat{m}(X)\right]-\beta\E_X\left[u(X)\Delta(\lambda,X)\right]} \\
            &\quad - \left(\E_X\left[m_{\lambda}(X)\right]\right)^{-1}\alpha\E_X\left[\left(1+\frac{m_{\lambda}(X)-\hat{m}(X)}{\hat{m}(X)\left(1-m_{\lambda}(X)\right)}\right)\Delta(\lambda,X)\right] =: A-B
        \end{aligned}
    \end{equation}
    where the third line follows from the fact that $g_{\lambda}(0,x)-\hat{g}(0,x)={\alpha v(x)}\Delta(\lambda,x)/{(1-m_{\lambda}(x))}$ and the fourth line from the fact that $\hat{m}(X)-\beta u(X)\Delta(\lambda,X) = m_{\lambda}(x)$, according to \eqref{eq:atte-construction}.
    
    Recall that $\theta_{\ml}^{\att} = {\E_X\left[\left(\hat{g}(1,X)-\hat{g}(0,X)\right)\hat{m}(X)\right]}/{\E_X\left[\hat{m}(X)\right]}$ and
    \begin{equation}
        \notag
        \E_X\left[u(X)\left(\hat{g}(1,X)-\hat{g}(0,X)\right)\Delta(\lambda,X)\right]-\theta_{\ml}^{\att} \E_X\left[u(X)\Delta(\lambda,X)\right] = 0
    \end{equation}
    by our choice of $u$ in Lemma \ref{lemma:choose-u}, so the first term $A$ in \eqref{eq:atte-expand} equals $\theta_{\ml}^{\att}$, since:
    \begin{align*}
        A =~& \frac{\theta_{\ml}^{\att} \E_X[\hat{m}(X)]-\beta\E_X\left[u(X)\left(\hat{g}(1,X)-\hat{g}(0,X)\right)\Delta(\lambda,X)\right]}{\E_X\left[\hat{m}(X)\right]-\beta\E_X\left[u(X)\Delta(\lambda,X)\right]}\\
        =~& \frac{\theta_{\ml}^{\att} \E_X[\hat{m}(X)]-\beta\theta_{\ml}^{\att} \E_X\left[u(X)\Delta(\lambda,X)\right]}{\E_X\left[\hat{m}(X)\right]-\beta\E_X\left[u(X)\Delta(\lambda,X)\right]}=\theta_{\ml}^{\att} 
    \end{align*}
    The second term can be further simplified as follows:
    \begin{subequations}
        \label{eq:atte-main-bound}
        \begin{align}
            B &= \left(\E_X\left[m_{\lambda}(X)\right]\right)^{-1}\alpha\E_X\left[\frac{m_{\lambda}(X)-\hat{m}(X)}{\hat{m}(X)\left(1-m_{\lambda}(X)\right)}\Delta(\lambda,X)\right]\nonumber \\
            &= -\alpha \beta \left(\E_X\left[m_{\lambda}(X)\right]\right)^{-1}\E_X\left[\frac{u(X)}{\hat{m}(X)\left(1-m_{\lambda}(X)\right)}\Delta(\lambda,X)^2\right]\nonumber \\
            &\leq -\alpha\beta\E_X\left[ \frac{u(X)}{\hat{m}(X)(1-m_{\lambda}(X))} \right] \label{eq:atte-main-bound-1} \\
            &= -\alpha\beta\E_X\left[ \frac{u(X)}{\hat{m}(X)(1-\hat{m}(X))} \right] - \alpha\beta \E_X\left[\frac{u(X)(m_{\lambda}(X)-\hat{m}(X))}{\hat{m}(X)(1-\hat{m}(X))(1-m_{\lambda}(X))} \right]\nonumber \\
            &= -\alpha\beta\E_X\left[ \frac{u(X)}{\hat{m}(X)(1-\hat{m}(X))} \right] + \alpha\beta^2 \E_X\left[\frac{u(X)^2\Delta(\lambda,X)}{\hat{m}(X)(1-\hat{m}(X))(1-m_{\lambda}(X))} \right]\nonumber \\
            &\leq -c_u\alpha\beta + 2c^{-3}C_u^2\alpha\beta^2 \leq -\frac{1}{2}c_u\alpha\beta \label{eq:atte-main-bound-2}
        \end{align}
    \end{subequations}
    where \eqref{eq:atte-main-bound-1} follows from $0<m_{\lambda}(X)<1$ and  $u(X)\geq 0$, and \eqref{eq:atte-main-bound-2} follows from $\left|m_{\lambda}(x)-\hat{m}(x)\right| \leq \beta C_u \leq c/2 \quad \Rightarrow\quad (1-m_{\lambda}(x))^{-1} \leq 2c^{-1}$
    and $\beta \leq c_uc^3C_u^{-2}/4$.
    Hence, for all $\lambda\in\{-1,+1\}^{M/2}$ we have $\theta_{\lambda}^{\att}\leq \theta_{\ml}^{\att}-c_u\alpha\beta/2$ as desired.
\end{proof}

We are now ready to prove Theorem 2. We choose $M$ sufficiently large according to Lemma \ref{lemma:atte-hellinger-bound}, $\gP=\{\hat{P}\}\cup\left\{Q_{\lambda}:\lambda\in\{0,1\}^{M/2}\right\}$, $P=\hat{P}$, $\pi$ be the discrete uniform distribution on  $\left\{Q_{\lambda}:\lambda\in\{0,1\}^{M/2}\right\}$, $s=c_u\alpha\beta/2=\Omega(\sqrt{e_nf_n})$ in the context of Lemma 2. Then all the listed conditions are satisfied for the ATE functional
\begin{equation}
    \notag
    T(P) = -\theta^{\att}(P) = -\E_P\left[ g(1,X)-g(0,X)\mid D=1\right].
\end{equation}
Therefore, by Lemma 2, we obtain a lower bound
\begin{equation}
    \notag
    \inf_{\hat{\theta}}\sup_{P\in\gP} \gQ_{P,1-\gamma}\left(\left| \hat{\theta}\left(\{(X_i,D_i,Y_i)\}_{i=1}^N\right) - \theta^{\att}\right|^2\right) = \Omega(\alpha^2\beta^2) = \Omega\left( e_n f_n\right).
\end{equation}

\section{Proof of the $\Omega(n^{-1})$ lower bounds}
\subsection{Proof of the $\Omega(n^{-1})$ lower bounds in Theorem \ref{thm:ate}}

We define
\label{subsec:proof-1/n}
\begin{equation}
    \label{eq:ate-1/n-construction}
    \begin{aligned}
        g(0,x) &= \hat{g}(0,x) \\
        g(1,x) &= \hat{g}(1,x) + \xi w(x) \\
        m(x) &= \hat{m}(x)
    \end{aligned}
\end{equation}
where $\xi$ is a constant that will be specified later.

Let $Q$ be the joint distribution of $(X,D,Y)$ induced by $g$ and $m$ defined above, then its density (w.r.t uniform measure) can be written as
\begin{equation}
    \notag
    q(x,d,y) = m(x)^d (1-m(x))^{1-d}g(d,x)^y(1-g(d,x))^{1-y}.
\end{equation}
From \eqref{eq:ate-1/n-construction} one can deduce that
\begin{equation}
    \notag
    \E_{X}\left[ w(x)\left( g(1,x)-g(0,x)\right) \right] = \E_{X}\left[ w(x)\left( \hat{g}(1,x)-\hat{g}(0,x)\right) \right] + \xi \Vert w\Vert_{P_X,2}^2
\end{equation}
and
\begin{equation}
    \notag
    \left| q(x,d,y)-\hat{p}(x,d,y)\right| \leq \xi |w(x)|.
\end{equation}
Moreover, by assumption we know that $\hat{p}(x,d,y) \geq c^2$, so we have that
\begin{equation}
    \notag
    H^2(\hat{P},Q) \lesssim \xi^2 \Vert w\Vert_{P_X,2}^2.
\end{equation}
By choosing $\xi\lesssim\frac{1}{\sqrt{n}\Vert w\Vert_{P_X,2}}$, one can guarantee that
\begin{equation}
    \notag
    H^2(\hat{P}^{\otimes n},Q^{\otimes n}) \leq n H^2(\hat{P},Q) \leq\alpha,
\end{equation}
so that the lower bound immediately follows from Theorem \ref{fano-method}.

\subsection{Proof of the $\Omega(n^{-1})$ lower bounds in Theorem \ref{thm:atte}} 

We consider the construction in \eqref{eq:ate-1/n-construction}. For the ATT, one can check that
\begin{equation}
    \notag
    \begin{aligned}
        \frac{\E_X\left[(g(1,X)-g(0,X))m(X) \right]}{\E_X[m(X)]} &= \frac{\E_X\left[(\hat{g}(1,X)-\hat{g}(0,X)+\xi)\hat{m}(X) \right]}{\E_X[\hat{m}(X)]} \\
        &= \frac{\E_X\left[(\hat{g}(1,X)-\hat{g}(0,X))\hat{m}(X) \right]}{\E_X[\hat{m}(X)]} + \xi.
    \end{aligned}
\end{equation}
The lower bound then directly follows from repeating the remaining steps in Section \ref{subsec:proof-1/n}.

\end{document}